\documentclass[journal]{IEEEtran}
\usepackage[linesnumbered, ruled,vlined]{algorithm2e}
\usepackage{times}
\usepackage{soul}
\usepackage{url}
\usepackage{graphicx}
\usepackage{amsmath}
\usepackage{booktabs}
\usepackage{nicematrix}
\usepackage{amsfonts}
\usepackage{array}
\usepackage{stfloats}
\usepackage{color}
\usepackage[tight,footnotesize]{subfigure}
\usepackage{adjustbox}
\usepackage{makecell}
\usepackage{caption}
\usepackage{epstopdf}
\usepackage{multirow}
\usepackage{multicol}
\usepackage{float}
\usepackage{comment}
\usepackage{orcidlink}

\begin{document}

\title{Target-specific Adaptation and Consistent Degradation Alignment for Cross-Domain Remaining Useful Life Prediction   
% Adversarial Domain Adaptation with Target Info Preservation and Degradation Stage Alignment for Machine Remaining Useful Life Prediction
}
%Time Series Unsupervised Domain Adaptation for Target-specific Information and Inconsistency Characteristics

% Enhancing Machine Remaining Useful Life Prediction: Domain Adaptation for Target-specific Insights and Inconsistency Characterization in Degradation Stage

%-------- double blind review --------%

\author{ Yubo Hou\,\orcidlink{0000-0002-0084-4380},
          Mohamed Ragab\,\orcidlink{0000-0002-2138-4395},
          Min Wu\,\orcidlink{0000-0003-0977-3600},~\IEEEmembership{Senior Member, IEEE, }
          Chee-Keong Kwoh\,\orcidlink{0000-0002-8547-6387},
          Xiaoli Li, \IEEEmembership{Fellow, IEEE, }
          Zhenghua Chen*, \IEEEmembership{Senior Member, IEEE}
 \thanks{Yubo Hou, Mohamed Ragab, Min Wu, Xiaoli Li and Zhenghua Chen are with Institute for Infocomm Research (I$^2$R), Agency for Science, Technology and Research (A*STAR), Singapore. (Email: yubo002@e.ntu.edu.sg, mohamedr002@e.ntu.edu.sg, wumin@i2r.a-star.edu.sg, xlli@i2r.a-star.edu.sg, chen0832@e.ntu.edu.sg)}
 \thanks{Yubo Hou, Chee-Keong Kwoh and Xiaoli Li are with School of Computer Science and Engineering, Nanyang Technological University, Singapore. (Email: asckkwoh@ntu.edu.sg)}
  \thanks{*Corresponding author: Zhenghua Chen}
 }

 %\thanks{E-mail: hou\_yubo@i2r.a-star.edu.sg, mohamedrma@i2r.a-star.edu.sg, chen0832@e.ntu.edu.sg, wumin@i2r.a-star.edu.sg, asckkwoh@ntu.edu.sg}
% The paper headers

\def\BibTeX{{\rm B\kern-.05em{\sc i\kern-.025em b}\kern-.08em
    T\kern-.1667em\lower.7ex\hbox{E}\kern-.125emX}}
\markboth{}
{Author \MakeLowercase{\textit{et al.}}: Title}
\maketitle

% As a general rule, do not put math, special symbols or citations
% in the abstract or keywords.

\begin{abstract}
% 1- Importance of RUL prediction task
Accurate prediction of the Remaining Useful Life (RUL) in machinery can significantly diminish maintenance costs, enhance equipment up-time, and mitigate adverse outcomes. Data-driven RUL prediction techniques have demonstrated commendable performance. However, their efficacy often relies on the assumption that training and testing data are drawn from the same distribution or domain, which does not hold in real industrial settings. To mitigate this domain discrepancy issue, prior adversarial domain adaptation methods focused on deriving domain-invariant features. Nevertheless, they overlook target-specific information and inconsistency characteristics pertinent to the degradation stages, resulting in suboptimal performance. To tackle these issues, we propose a novel domain adaptation approach for cross-domain RUL prediction named TACDA. Specifically, we propose a target domain reconstruction strategy within the adversarial adaptation process, thereby retaining target-specific information while learning domain-invariant features. Furthermore, we develop a novel clustering and pairing strategy for consistent alignment between similar degradation stages. Through extensive experiments, our results demonstrate the remarkable performance of our proposed TACDA method, surpassing state-of-the-art approaches with regard to two different evaluation metrics. 
Our code is available at \url{https://github.com/keyplay/TACDA}.

% We propose a novel domain adaptation approach for cross-domain RUL prediction. Our proposed method is founded upon an adversarial domain adaptation architecture while leveraging a reconstruction loss, allowing it to incorporate target-specific information while learning domain-invariant features effectively. Additionally, we introduce a data selection mechanism to align data within the same degradation stage.
% 7- Experiments and results

\end{abstract}
%Specifically, we utilize discriminator to make target domain features similar to source domain features. To preserve the target domain information, we utilize decoder to reconstruct target data from target domain features based on dynamic time warping. In data selection mechanism, we group source data based on health index and RUL label and group target data based on clustering method. Then we match source and target group with similar entropy for domain adaptation.

\renewcommand{\abstractname}{Note to Practitioners}
\begin{abstract}
This paper introduces TACDA, a deep neural network crafted for domain adaptation, aimed at the task of predicting the remaining useful life (RUL) of machinery using time series data from multiple sensors under different operating conditions. In particular, it utilizes limited labeled data in one operating condition (source domain) and unlabeled data in another condition (target domain) to train a model. This well-trained model is capable of predicting the RUL in the target domain based on sensor data. Unlike existing methods, TACDA can maintain information specific to the target domain and align the data within the same degradation stage for better adaptation performance. 
A current limitation of TACDA is its need for simultaneous access to data from both the source and target domains for training. Future research will explore source-free domain adaptation, which relies solely on unlabeled data from the target domain and a model initially trained in the source domain.
\end{abstract}


\IEEEpeerreviewmaketitle

\section{Introduction}
\label{introduction}
% Prognostic and health management 
%Prognostic and health management (PHM) of industrial systems and equipment can enhance reliability, reduce maintenance costs, and enhance safety and operational performance \cite{kordestani2023overview}. In the domain of PHM, predicting the remaining useful life (RUL) is a pivotal task for well-informed maintenance decisions.
Prognostic and Health Management (PHM) of industrial systems and equipment plays a crucial role across diverse industries, including manufacturing, aerospace, and energy. PHM helps detect potential failures and forecast the lifespan of critical components, ultimately improving operational safety and efficiency. In aerospace, for instance, a significant percentage of aircraft breakdowns are due to turbofan engine failures, underscoring the urgent need for advanced PHM solutions that can prevent catastrophic incidents and reduce costly unscheduled maintenance. Similarly, in energy production, unplanned downtime due to equipment failure can lead to substantial financial losses and disrupt energy supply, emphasizing the value of PHM in ensuring consistent operations. By leveraging PHM, organizations can enhance equipment reliability, reduce maintenance costs, and optimize performance, ultimately minimizing the risk of unexpected failures~\cite{kordestani2023overview}. Within the domain of PHM, accurately predicting the Remaining Useful Life (RUL) of machinery is pivotal, as it enables proactive maintenance decisions.
% Prognostic and health management (PHM) is a field that focuses on the prediction, diagnosis, and management of the health condition of industrial systems and equipment, e.g., manufacturing machines and aero-engines. The PHM technology can improve reliability, reduce maintenance costs, optimize maintenance schedules, and enhance safety and operational performance \cite{hu2022prognostics}. A critical aspect of PHM is the remaining useful life (RUL) prediction of an equipment, as it enables predictive maintenance and assist of the maintenance decisions. RUL refers to the estimated remaining operational time of a system before it reaches failure. 
% approach for RUL: model-based, data-driven 
Various approaches have been proposed for RUL prediction, which can generally be classified into three categories: model-based approaches, data-driven approaches, and hybrid approaches. Specifically, model-based approaches focus on describing the degradation behavior of a system using a mathematical or physics model, which requires a strong theoretical understanding \cite{wang2020theoretical, liu2013data}. As the mechanical structure becomes increasingly complex, predicting RUL with model-based approaches can be extremely challenging. Meanwhile, with the increasing availability of data from deployed sensors, data-driven approaches become popular for RUL prediction~\cite{wang2023local, behera2023gan, wang2023joint, nejjar2024domain}. Hybrid approaches utilize both the pre-established model of model-based approaches and prediction ability of data-driven approaches based on historical data, which can improve the accuracy of prognostic result.
For example, \cite{chang2017new} integrates an unscented Kalman filter with complete ensemble empirical mode decomposition and a relevance vector machine, while \cite{chang2019hybrid} couple a particle filter with a relevance vector machine.

% Data-driven approaches rely on collected data to learn patterns between the operational conditions and the degradation process, while it does not require expert knowledge of the degradation mechanism.

%Given approaches encompass machine learning algorithms, e.g., bayesian networks~\cite{camci2010health}, extreme learning machine~\cite{skordilis2019double}, support vector regression (SVR)~\cite{soualhi2014bearing}, and random forest~\cite{alfarizi2022optimized}, as well as deep learning methods, including long short-term memory (LSTM) networks~\cite{behera2023gan, chen2020machine}, convolutional neural network (CNN)~\cite{liu2019simultaneous}, and hybrid model~\cite{yang2020efficient}. 
Despite the promise of data-driven approaches, their success is mainly attributed to two main assumptions, i.e., sufficient labeled data and independent and identically distributed (IID) data~\cite{da2020remaining}. However, these assumptions are often impractical in real industrial settings for the subsequent reasons. First, the acquisition of annotated data (failures) incurs significant costs. Additionally, the degradation process of machines can span prolonged periods, further constraining the accessibility of faulty data~\cite{lei2018machinery}. Secondly, considering the dynamics of a real environment, a model is usually trained under one operating condition and tested under different operating conditions, leading to a significant performance deterioration due to the domain shift problem \cite{pan2010survey}.

Given the aforementioned challenges, accurately predicting the RUL for machines under various working conditions with limited labeled data poses significant difficulties. Unsupervised domain adaptation (UDA) is a promising technique that facilitates knowledge transfer from a labeled source domain to a distinct yet related unlabeled target domain~\cite{pan2010survey}. Recently, there has been an increased focus on UDA for RUL prediction task. Predominantly, the existing methods endeavor to identify domain-invariant features to mitigate the distribution shift problem, either through adversarial training~\cite{ragab2020adversarial, jiang2022adversarial, da2020remaining} or by minimizing the statistical distance between domains~\cite{jia2017assessment, cheng2021transferable, mao2019predicting}.
% The mainstream UDA approach is to acquire domain-invariant features based on two main principles: metric-based approaches~\cite{jia2017assessment, cheng2021transferable, mao2019predicting} and adversarial-based approaches~\cite{ragab2020adversarial, jiang2022adversarial, da2020remaining}. 
Despite the promise of these methods in addressing the domain shift problem with scarce target labels,  they still exhibit the following limitations.
% These approaches have exhibited commendable performance in RUL prediction and demonstrated the efficacy of UDA in mitigating the scarcy of labeled data of the target domain. However, they still present the following limitations.
\begin{itemize}
    \item \textbf{Lack of target-specific information}: Solely pursuing domain-invariant features does not assure optimal performance in the target domain~\cite{stojanov2021domain, ragab2020contrastive}. Enforcing target domain features to globally resemble the source domain features without appropriate constraints may lead to the removal of target-specific information.
    \item \textbf{Inconsistent characteristics for degradation stages}: RUL degradation can be categorized into distinct stages based on degradation speed~\cite{liu2021prediction}. Each stage possesses unique characteristics. Existing methods bring the two domains closer globally, which may blend samples from different degradation stages~\cite{li2018domain, chen2019progressive}. For instance, early-stage degradation in the target domain might be aligned with late-stage degradation in the source domain, resulting in misalignment.
\end{itemize}

\begin{comment}
\begin{figure}[]
\centering
 \includegraphics[width=0.45\textwidth]{images/problem_intro2.pdf}
\caption{Comparisons between previous UDA methods and ours.}
\label{fig:UDA2}
\end{figure}
\end{comment}

\begin{figure}[]
\centering
 \includegraphics[width=0.47\textwidth]{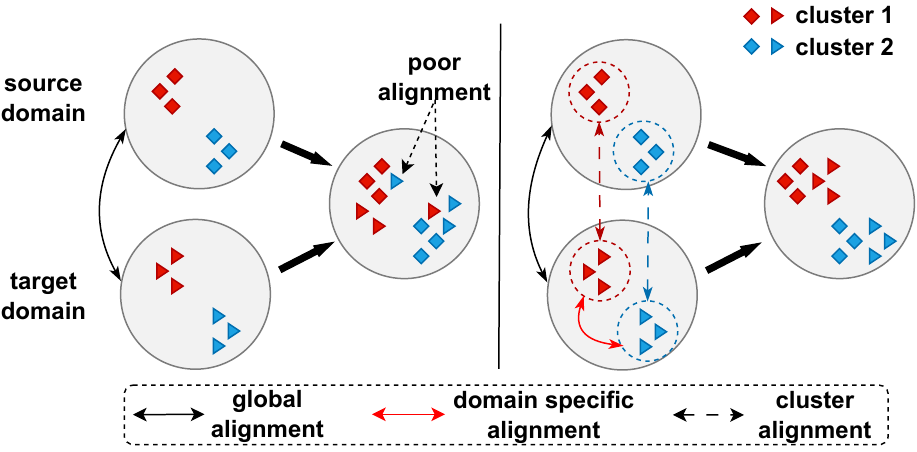}
\caption{Comparisons between previous UDA methods and ours. Left: only pursuing global alignment results in poor alignment. Right: considering target domain-specific information and inconsistency characteristics of clusters reduces the feature discrepancy.}
\label{fig:UDA}
\end{figure}
% our proposed work
As presented in Fig.~\ref{fig:UDA} (left), the above-mentioned limitations could potentially hinder the overall model performance. In response to these challenges, we propose a novel target-specific adaptation and consistent degradation alignment (TACDA) approach for machine RUL prediction across different working conditions. 
Specifically, to tackle the lack of target-specific information issue, %we employ a soft-DTW~\cite{cuturi2017soft} based auto-encoder to preserve target domain information during the domain adaptation process for time series data.
we design an auto-encoder based domain alignment module with a soft dynamic time warping (soft-DTW) loss to preserve target domain information for time series sensory data. To address the ignore of inconsistency characteristics for the degradation stage problem, we not only make domain alignment on global data but also make degradation alignment between source and target domains. Through extensive experimentation, we have thoroughly evaluated the performance of our proposed TACDA method in accurately predicting the RUL of machines across diverse operating conditions.

%We propose a novel approach that employs a soft-DTW~\cite{cuturi2017soft} based auto-encoder to preserve target domain information during the domain adaptation process for time series data. We undertake joint optimization of the target feature extractor to minimize both the domain adaptation loss and the reconstruction loss. Specifically, the domain adaptation loss guides the target feature extractor to generate features resembling those of the source domain, while the reconstruction loss preserves target-specific information through the reconstruction of target data. Moreover, we not only make domain alignment on global data, but also make domain alignment on data in the same degradation stage. We have devised a data selection mechanism that effectively segregates the source data into different stages based on the RUL label and utilizes soft-DTW based K-means clustering and SVD entropy to partition the target data. Through extensive experimentation, we have thoroughly evaluated the performance of our proposed TACDA method in accurately predicting the RUL of machines across diverse working conditions.

The main contributions of this study are listed as follows.
\begin{itemize}
     \item We propose an innovative approach called TACDA, which considers both the target-specific adaptation and consistent degradation alignment for RUL prediction.
    \item  We leverage the auto-encoder to preserve target-specific information, while learning invariant representations. In particular, we employ soft-DTW instead of mean squared error as reconstruction loss within the auto-encoder to preserve target-specific information. 
    \item We propose a clustering and pairing strategy, which clusters the data into different degradation stages and then aligns the data within the same degradation stage for better adaptation performance. 
\end{itemize}

\begin{comment}
\begin{itemize}
    \item We have developed an innovative approach called the target-specific adaptation and consistent degradation alignment (TACDA) method, specifically designed to tackle the demanding yet pragmatic task of machine RUL prediction. Our approach effectively transfers valuable knowledge for RUL prediction from one condition (distribution/domain) to another, demonstrating its potential in bridging the gap between different operational scenarios.
    \item  We have proposed an innovative solution that leverages the soft-DTW based auto-encoder to learn an invariant representation and maintain the target domain information. By doing so, we have demonstrated that this approach yields satisfactory performance in predicting the RUL.
    \item We have devised a data selection mechanism that focuses on aligning data within the same degradation stage. This approach facilitates domain alignment and enables us to achieve further improvements in performance.
\end{itemize}
\end{comment}
\section{Related works}
\label{related_works} 

\subsection{Deep Learning for RUL Prediction}
Deep learning for RUL prediction is broadly categorized into two types: feed-forward neural networks and recurrent neural networks (RNNs). In~\cite{zhu2018estimation}, convolutional neural networks (CNNs) was utilized to extract features at multi-scale for detecting fault growth and predicting machine RUL. In~\cite{liu2019simultaneous}, a CNN network was proposed with a joint loss function to simultaneously identify faults and provide predictions for RUL. Another study~\cite{deutsch2017using} employed a Deep Belief Network (DBN) for feature extraction and RUL prediction. 

For dynamic systems, RNNs offer sequential modeling capabilities. Among RNN architectures, long short-term memory (LSTM) networks have gained significant popularity due to their ability to capture long-term dependencies and overcome the vanishing gradient problem associated with traditional RNNs. In~\cite{huang2019bidirectional}, a bidirectional LSTM (BiLSTM) methodology was introduced, which is enhanced with auxiliary features to forecast the RUL across diverse operational conditions. The bidirectional nature of BiLSTM allows it to capture dependencies from both past and future time steps. In~\cite{chen2020machine}, an attention-based LSTM approach was developed for RUL prediction. This method dynamically selects important features using attention mechanisms, resulting in accurate RUL predictions. However, these methods cannot work well with insufficient labeled data and non independent and identically distributed data in real world application.

\subsection{Unsupervised Domain Adaptation for RUL Prediction}
Unsupervised Domain Adaptation (UDA) for RUL prediction aims to reduce labeling costs by training neural networks to transfer knowledge from a labeled source domain to an unlabeled target domain. Existing UDA methods focus on minimizing domain discrepancy to achieve high performance on the target domain. These methods can be categorized into two branches: metric-based and adversarial-based methods.

%Metric-based methods use statistical distances to align the feature representations of the source and target domains. Deep domain confusion (DDC)~\cite{jia2017assessment} and AdvSKM~\cite{liu2021adversarial} employ the maximum mean discrepancy (MMD) to address the challenge of domain discrepancy. TCNN~\cite{cheng2021transferable} incorporated a multiple-kernel MMD to reduce distribution discrepancy between the source and target domains. Correlation alignment (CORAL)~\cite{sun2017correlation} focused on minimizing the covariance shift between the feature distributions of the source and target domains. In~\cite{mao2019predicting}, a solution was exploited by combining contractive denoising autoencoder and transfer component analysis (TCA). 

Metric-based methods align the feature representations of source and target domains by leveraging statistical distances. Deep Domain Confusion (DDC)~\cite{jia2017assessment}, AdvSKM~\cite{liu2021adversarial} and FDMTCN~\cite{yan2024feature} utilize the maximum mean discrepancy (MMD) to mitigate domain differences. To address distributional shifts, TCNN~\cite{cheng2021transferable} incorporates a multi-kernel MMD approach for better alignment between source and target domains. Correlation alignment (CORAL)~\cite{sun2017correlation} aims to reduce the covariance shift between feature distributions across domains. Additionally, in~\cite{mao2019predicting}, a hybrid strategy combining the contractive denoising autoencoder and transfer component analysis (TCA) was implemented.

%Adversarial-based methods employ domain discriminator networks to compel the feature extractor to acquire representations that are invariant across domains. Domain adversarial neural network~\cite{da2020remaining} utilized a reverse gradient strategy to conduct adversarial training for both the domain classifier and the feature extractor. Adversarial domain adaption approach for remaining useful life prediction (ADARUL)~\cite{ragab2020adversarial} utilized a conventional GAN loss with flipped labels to learn domain-invariant features. 
Adversarial-based methods use domain discriminator networks to encourage the feature extractor to learn domain-invariant representations. The domain adversarial neural network (DANN)~\cite{da2020remaining, deng2023calibration} employs a gradient reversal strategy to facilitate adversarial training between the domain classifier and the feature extractor. Similarly, the adversarial domain adaptation for remaining useful life prediction (ADARUL)~\cite{ragab2020adversarial} adopts a traditional GAN loss with label flipping to achieve domain-invariant feature learning. Adversarial regressive domain adaptation approach~\cite{jiang2022adversarial} used discriminator and bi-regressors to align distributions for infrared thermography-based RUL prediction. In~\cite{li2022domain}, a transformer-based method was proposed which leverages feature-level and semantic-level discriminators for distribution alignment. In~\cite{chen2023aero}, a method was proposed that integrates clustering for sensor selection and ensemble technology for transfer learning. However, these methods do not incorporate target-specific information during the process of learning domain-invariant features, which harms the performance in target domain.
In~\cite{nejjar2024domain}, a method was proposed that leverages an adversarial framework that separately aligns the marginal distributions of different operation phases between source and target domains.
%Conditional domain adversarial network~\cite{long2018conditional} incorporated the task-knowledge with features during the domain alignment phase.

%In the realm of Remaining Useful Life (RUL) prediction, limited efforts have been made to tackle the challenge of knowledge transfer across distinct domains. In~\cite{da2020remaining}, a deep domain adaptation (DDA) technique was proposed for RUL prediction, leveraging unlabeled data from the target domain. The DDA approach employed an LSTM network to extract features and incorporated the reverse gradient technique to mitigate the domain shift issue. In a recent development, Contrastive adversarial domain adaptation (CADA)~\cite{ragab2020contrastive} is constructed upon an adversarial domain adaptation framework, incorporating a contrastive loss. This design enables the model to incorporate target-specific information during the process of learning domain-invariant features. In~\cite{chen2023aero}, a method was proposed that integrates clustering for sensor selection and ensemble technology for transfer learning.

\section{Methodology}
\label{sec:methodology}

\begin{figure*}[]
\centering
 \includegraphics[width=1\textwidth]{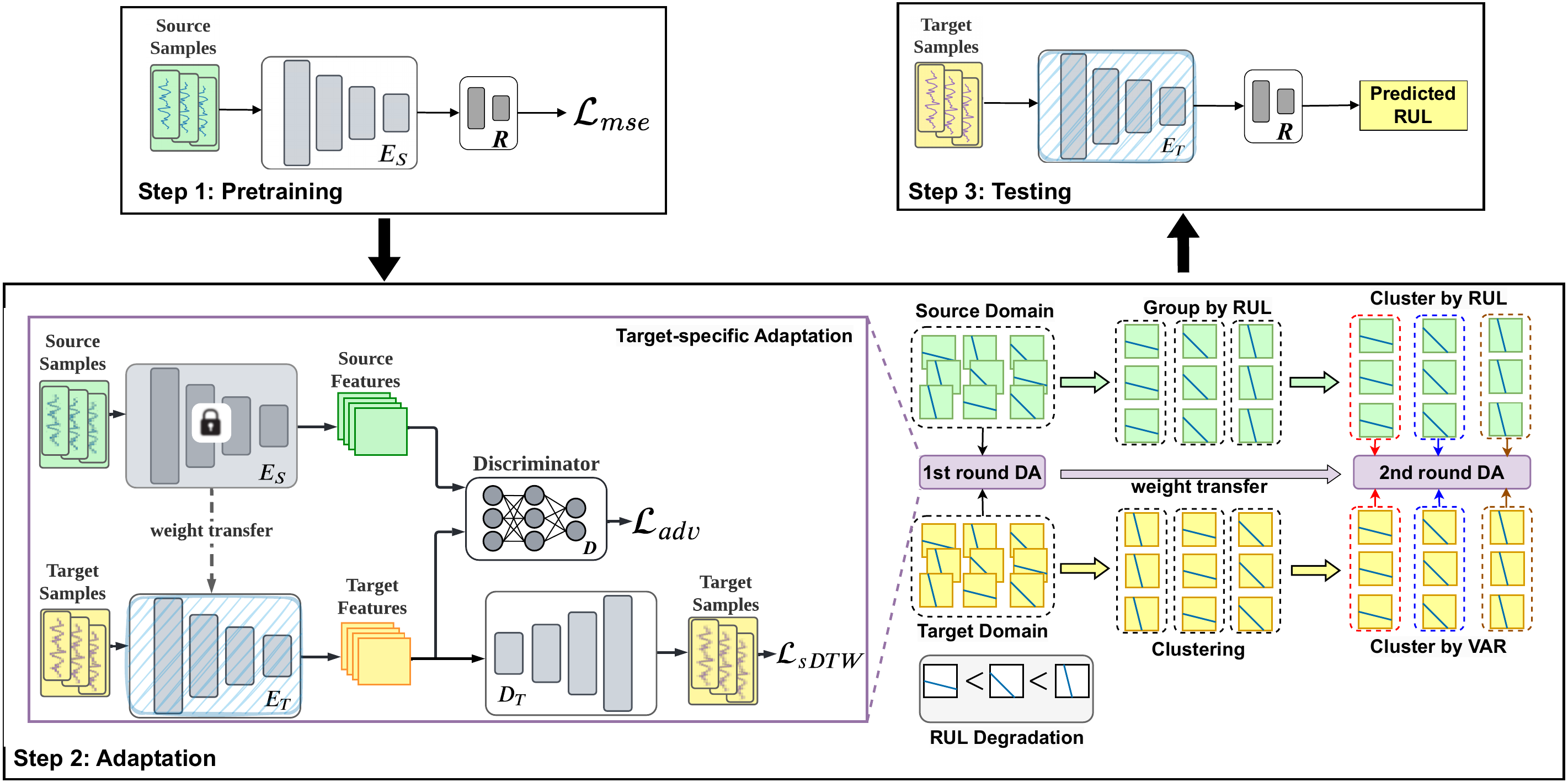}
\caption{Our proposed TACDA approach. Target-specific Adaptation (TA) learns the target encoder $E_T$, the target decoder $D_T$ and discriminator $D$ via optimizing the adversarial loss $\mathcal{L}_{adv}$ and the soft-DTW loss $\mathcal{L}_{sDTW}$. Specifically, 
%TA is used in TACDA twice, namely,  all the source and target data are fed into TA in the 1st round DA, while the clustered and paired source and target data are fed into TA in the 2nd round DA. 
TA is applied in two rounds:  
1st Round DA: All source and target data are fed into TA for global domain alignment.  
2nd Round DA: Clustered and paired source-target data (grouped by degradation stages, e.g., stage 1, 2, 3) are iteratively aligned in a loop (stage 1 → stage 2 → stage 3) for stage alignment.  
 After the 2nd round DA, the optimized $E_T$ is directly transferred to Step 3.}
\label{fig:AAEDA}
\end{figure*}

\subsection{Problem Formulation}
We denote a source domain with $N_S$ labeled samples $\{X^i_S, y^i_S\}^{N_S}_{i=1}$ and a target domain with $N_T$ unlabeled samples $\{X^i_T\}^{N_T}_{i=1}$, where $X^i_S \in \mathbb{R}^{M \times L}$ and $X^i_T \in \mathbb{R}^{M \times L}$ are both multivariate time series data consisting of $M$ sensors and $L$ time steps. $y^i_S$ is the RUL label. We aim to transfer knowledge from labeled source domain to unlabeled target domain and then improve the performance of RUL prediction on the target. Table \ref{tab:notation} provides a summary of the notations employed in this paper.

\begin{table}[]
\centering
\caption{List of notations.}
\label{tab:notation}
\begin{tabular}{ll}
\hline
Notation & Definition                      \\ \hline
 $X_S/X_T$        & source/target data              \\ \hline
 $y_S$        & source RUL label              \\ \hline
 $N_S/N_T$       & number of source/target samples \\ \hline
 $f_S/f_T$        & source/target features          \\ \hline
 $E_S/E_T$        & source/target encoder           \\ \hline
 $D_T$        & target decoder                  \\ \hline
 $D$        & domain discriminator            \\ \hline
 $R$        & predictor            \\ \hline
 $M$        & number of sensors               \\ \hline
 $L$        & sequence length                 \\ \hline
\end{tabular}
\end{table}

\subsection{Overview}

\begin{comment}
\begin{figure}[t]
\centering
 \includegraphics[width=0.45\textwidth]{images/AAEDA_flow.pdf}
\caption{Flowchart of the proposed AAEDA approach. \textcolor{red}{remove}}
\label{fig:overview}
\end{figure}
\end{comment}

The overall structure of our proposed TACDA method is shown in Fig.~\ref{fig:AAEDA}. TACDA consists of three main components: source encoder $E_S$, target encoder $E_T$, and shared predictor $R$. 
First, we pretrain the source encoder $E_S$ and predictor $R$ to learn the RUL distribution from the source domain. 
Then, in the adaptation phase, we train the target encoder $E_T$. This adaptation phase is divided into two stages. In the first stage, both source and target data are fed to the target-specific adaptation (TA) module to preserve target-specific information while learning invariant representations. To ensure consistent degradation alignment, in the second stage, source and target data are clustered, respectively. Then, data groups from source and target domains are paired by similar status. In the end, multiple paired source and target data are fed into the TA module successively for fine-tuning the model parameters from the first stage. 
Once training is complete, we use the well-trained target encoder $E_T$ and predictor $R$ to estimate RUL in the target domain.
The following sections will introduce each module in detail.

\subsection{Pretraining on Source Domain}
During this phase, the objective is to train a model using the labeled source domain data to have the knowledge about the RUL prediction task. Specifically, a source encoder $E_S$ and RUL predictor $R$ are trained on source data $X_S$. The source encoder extracts features from source data: $f_S=E_S(X_S)$. The RUL predictor estimates RUL value according to the extracted features. The source encoder and the RUL predictor are optimized using the mean square error loss:
\begin{equation}
L_{mse} = \frac{1}{N_S}\sum_{i=1}^{N_S} (y_i-\widehat{y_i})^2,
\label{eq:mse_loss}
\end{equation}
where $\widehat{y_i} = R(E_S(X_S))$ is the estimated RUL value, $y_i$ is the true RUL label, and $N_S$ represents the number of samples in source domain.

\subsection{Target-specific Adaptation}
Given the pretrained source model, the focus lies in achieving adaptation upon the unlabeled target data. To do so, the target-specific adaptation module is proposed, containing adversarial adaptation for domain invariant feature and the decoder for preserving target-specific information. The target-specific adaptation module, depicted in Fig.~\ref{fig:AAEDA} (left), comprises a trained source encoder $E_S$, a target encoder $E_T$, a target decoder $D_T$ and a domain discriminator $D$. The $E_T$ is initialized with the weights of $E_S$ which was pre-trained on labeled source data $\{X_S, y_S\}$. The source domain feature $f_S$ and target domain features $f_T$ are extracted from source data $X_S$ and target $X_T$ via $E_S$ and $E_T$, respectively. Then extracted features $f_S$ and $f_T$ are fed into the domain discriminator $D$ to minimize their discrepancy. Simultaneously, the target features $f_T$ are passed through the decoder $D_T$ to get reconstructed target data $X'_T$. The difference between $X_T$ and $X'_T$ is computed to preserve target-specific information during alignment. The formalized procedure of the target-specific adaptation module in our approach is outlined in Algorithm \ref{al:AAEDA}. The detailed procedures are described in the subsequent paragraphs.

% \begin{figure*}[t]
% \centering
%  \includegraphics[width=1\textwidth]{images/AEDA_RUL.pdf}
% \caption{Proposed TACDA approach.}
% \label{fig:AAEDA}
% \end{figure*}

\subsubsection{Adversarial Adaptation}

The training procedure of adversarial adaptation between the domain discriminator $D$ and the target encoder $E_T$ can be defined as follows:
\begin{align}
\min_{E_T}\max_{D}\mathcal{L}_{adv}= &\mathbb{E}_{X_S\sim P_S} \left[ log D(E_S(X_S)) \right] \nonumber \\ + ~ &\mathbb{E}_{X_T\sim P_T} \left[ log(1-D(E_T(X_T))) \right].
\label{eq:adv}
\end{align}
The target encoder $E_T$ is trained to minimize $\mathcal{L}_{adv}$, while the discriminator $D$ is updated in an adversarial manner to maximize $\mathcal{L}_{adv}$. Consequently, the trained target encoder $E_T$ becomes capable of extracting features $f_T$ that have minimum disparity compared to the source features $f_S$.

\subsubsection{Target-specific Decoder}
Adversarial domain adaptation demonstrates efficacy in learning target domain features that exhibit invariance with respect to the source features. Nevertheless, in its pursuit to minimize the adversarial loss, it runs the risk of removing task-specific information from the target features, thereby potentially compromising performance in the target domain. Consequently, it is imperative to preserve target-specific information throughout the process of domain adaptation. 
%it may remove task-specific information from the target features to minimize the adversarial loss, which can deteriorate the performance on the target domain. Hence, it is required to preserve target-specific features during the domain alignment task. 
To achieve this objective, we develop an auto-encoder based on soft-DTW loss~\cite{cuturi2017soft}, which is robust to the shifts and dilatation across the time dimension compared with conventional MSE loss. 

\begin{figure}[]
\centering
 \includegraphics[width=0.35\textwidth]{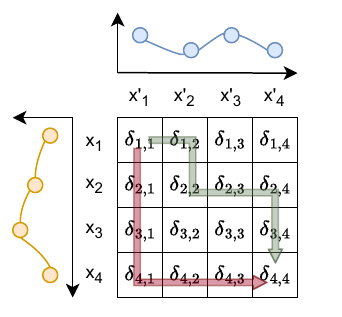}
\caption{Two warping paths (red and green, as well as the top-left and bottom-right entries) between two time series of length 4. The cost of a warping path is determined by the cumulative sum of difference $\delta$ visited along the path.}
\label{fig:dtw}
\vspace{-2ex}
\end{figure}

Soft-DTW measures the similarity between $X_T$ and $X'_T$ by considering all possible warping path mapping the elements of $X_T$ to $X'_T$. Fig.~\ref{fig:dtw} shows an example of warping path between two time series of length 4. Soft-DTW can be expressed as follows:

\begin{equation}
\mathcal{L}_{sDTW} = -\gamma log \sum_{A \in \mathcal{A}_{L, L}} e^{-\bigl \langle A, \triangle(X_{T}, X'_{T}) \bigr \rangle / \gamma}, 
\label{eq:softdtw}
\end{equation}
where $\gamma$ is a smoothing parameter, $\triangle(X_{T}, X'_{T})=[\delta(x_i, x'_j)]_{ij}\in \mathbb{R}^{L\times L}$, $\delta(x_i, x'_j)$ is the quadratic Euclidean distance and $\bigl \langle A, \triangle(X_{T}, X'_{T}) \bigr \rangle$ is inner product. The set $\mathcal{A}_{L, L} \subset \{0,1\}^{L\times L}$ represents the collection of (binary) warping paths on an $L\times L$ matrix. These paths connect the upper-left matrix entry (1, 1) to the lower-right one ($L, L$) using exclusively $\downarrow$, $\rightarrow$, and $\searrow$ moves. %$\mathcal{A}_{L, L} \subset \{0,1\}^{L\times L}$ for the set of (binary) warping path is paths on a $L\times L$ matrix that connect the upper-left (1, 1) matrix entry to the lower-right ($L, L$) one using only $\downarrow$, $\rightarrow$, $\searrow$ moves.

\begin{algorithm}[h]
\DontPrintSemicolon
\KwIn{Source domain: $\{X_S, y_S\}$,
      Target domain: $X_T$}
\KwOut{Trained target encoder $E_T$, Trained target decoder $D_T$, Trained discriminator $D$}
$E_S \gets$ Pretrained source encoder\;
$E_T \gets$ Initialized with $E_S$ parameters\;
$D_T \gets$ Target decoder\;
$D \gets$ Domain discriminator\;
\For{number of iterations}{
  $f_S \gets E_S(X_S)$ \tcp{Extract source features}
  $f_T \gets E_T(X_T)$ \tcp{Extract target features}
  Feed $f_S$ and $f_T$ to $D$\;
  Calculate adversarial loss $\mathcal{L}_{adv}$ using Eq.~(\ref{eq:adv})\;
  Update $D$ by $\mathcal{L}_{adv}$\;
  $X'_T \gets D_T(f_T)$ \tcp{Reconstruct target sample}
  Calculate soft-DTW loss $\mathcal{L}_{sDTW}$ using Eq.~(\ref{eq:softdtw})\;
  Update $D_T$ by $\mathcal{L}_{sDTW}$\;
  Update $E_T$ by $\mathcal{L}_{adv}+\lambda\mathcal{L}_{sDTW}$\;
}
\Return{$E_T$, $D_T$, $D$}
\caption{Target-specific Adaptation (TA)}
\label{al:AAEDA}
\end{algorithm}

\subsection{Consistent Degradation Alignment}

The target-specific adaptation module aligns the distributions of source and target data on a global scale. However, it overlooks discrepancies in characteristics across different degradation stages, leading to suboptimal alignment. To address this issue, we propose a consistent degradation alignment module to ensure alignment between analogous degradation stages during the adaptation phase.  Initially, we classify the degradation of both source and target domains into three stages based on their degradation speed: sluggish, accelerated, and terminal. %, as in \cite{liu2021prediction}.

\subsubsection{Identifying Source Degradation Stages}

\begin{figure}[]
\centering
 \includegraphics[width=0.5\textwidth]{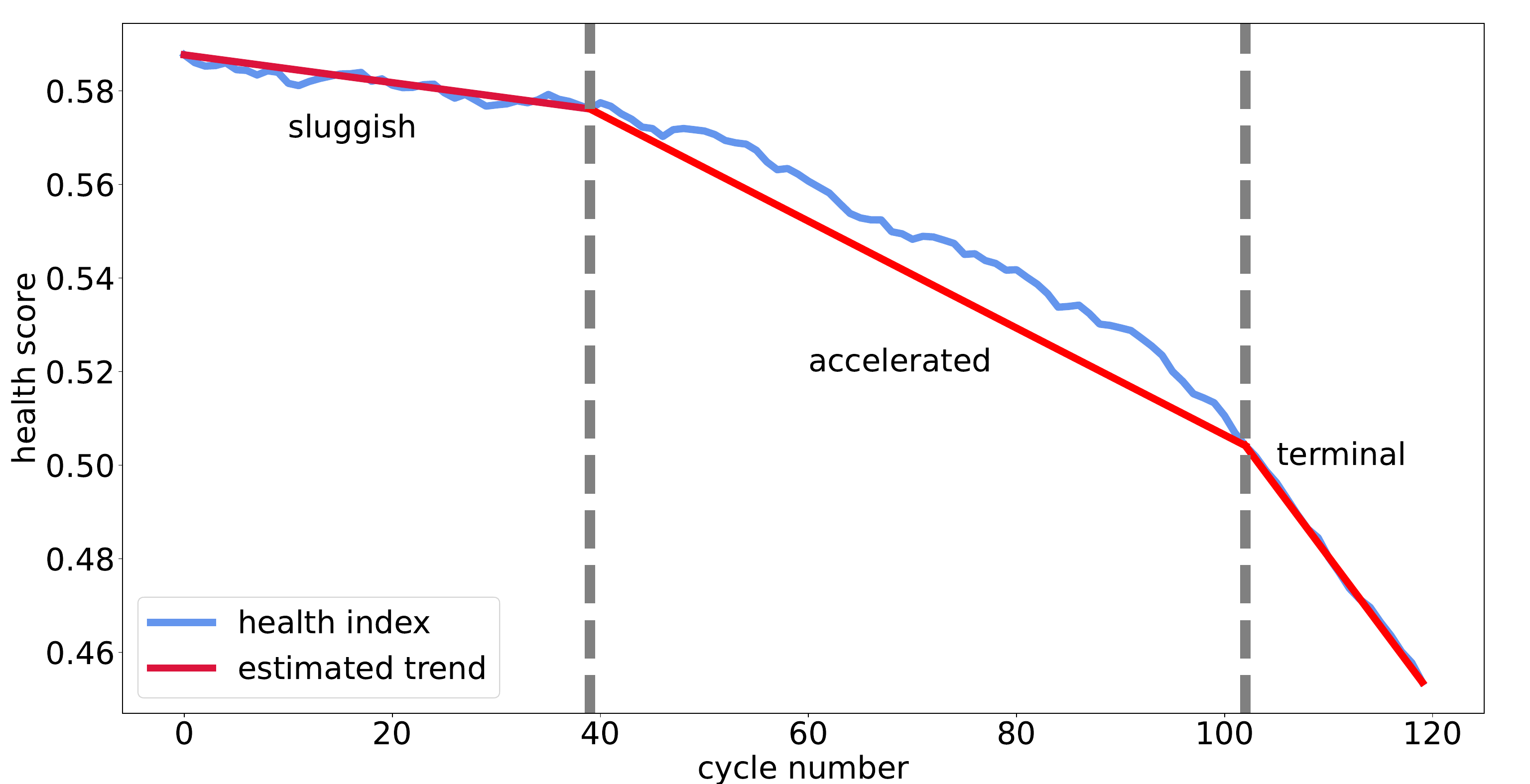}
\caption{Three degradation stages categorized by the health index.}
\label{fig:health_idx}
\end{figure}

\begin{figure}[]
\centering
 \includegraphics[width=0.5\textwidth]{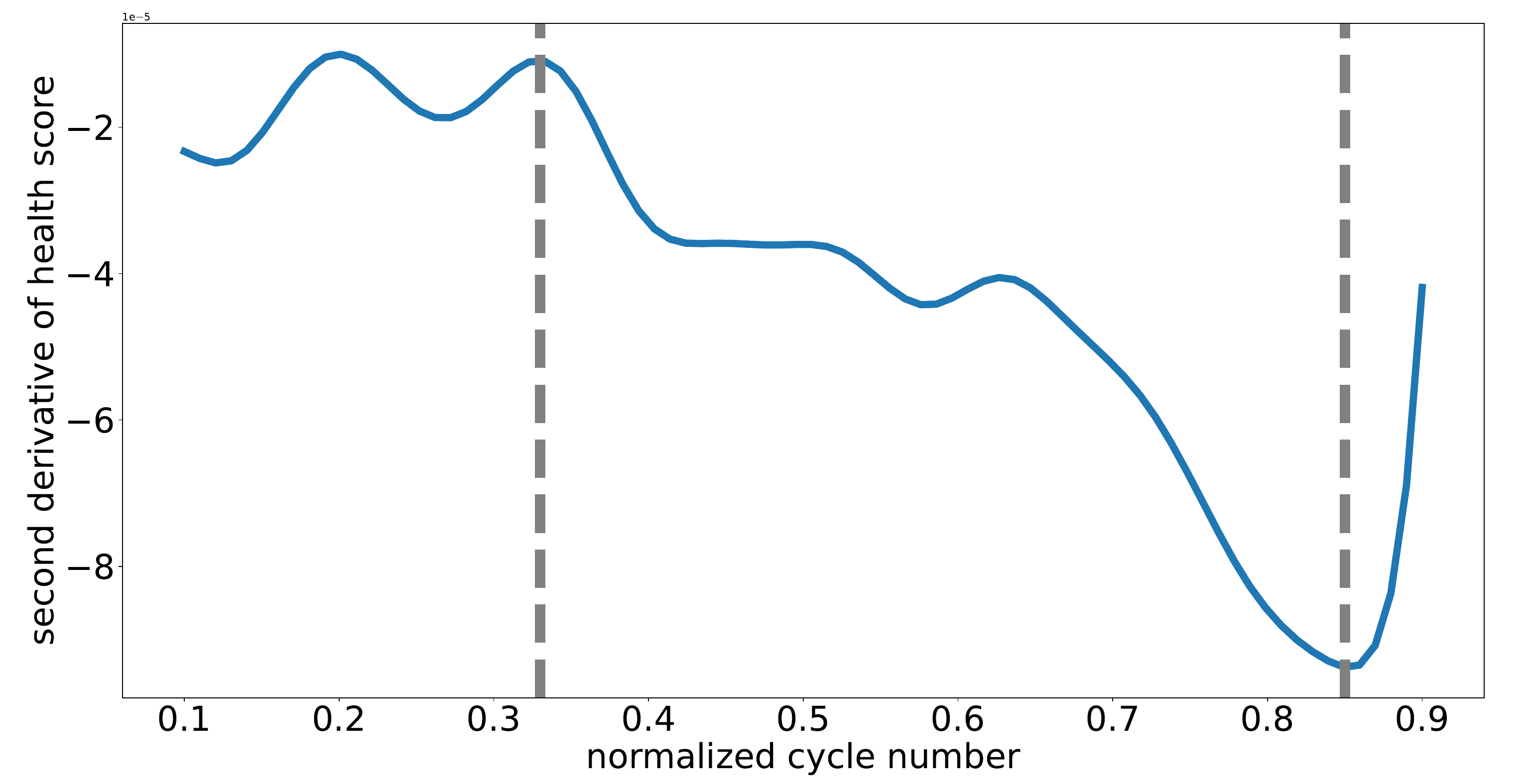}
\caption{Second derivative of health index.}
\label{fig:2nd_dev_health_idx}
\end{figure}

For the source domain data, we partition each engine's lifecycle into three degradation stages using available labels. To define precise stage boundaries, we compute a health index (HI) through a linear combination of critical sensor readings, following methodologies in \cite{lei2018machinery, liu2021prediction}.
As illustrated in Fig.~\ref{fig:health_idx}, which plots the HI trajectory of a representative engine against operational cycles, this index quantifies progressive degradation by synthesizing multi-sensor signals.
The HI trajectory is analyzed in conjunction with RUL labels to identify distinct degradation phases. The second derivative of the averaged HI reflects how the rate of health degradation (first derivative) changes over normalized cycle numbers.

As outlined in Algorithm \ref{al:aHI}, we begin by discretizing the normalized RUL values \(y_i \in [0,1]\) into \(K\) equally spaced bins. Each bin \(k\) accumulates the health‐index observations \(h_i\) whose corresponding \(y_i\) falls into that bin, and we track the count of samples per bin to enable averaging. After iterating over all \(N\) samples, we obtain an average health index \(\overline{h}_k\) at each bin.
To reduce noise and highlight meaningful curvature in the degradation profile, we apply a Gaussian smoothing filter to the sequence \(\{\overline{h}_k\}_{k=0}^K\). Finally, we approximate the second derivative of HI.
In our experiments, we set \(K = 100\).

\begin{algorithm}[h]
\DontPrintSemicolon
\KwIn{RUL label: $y \in [0,1]$,
      Health index: $h$}
\KwOut{Second Derivative of HI: $\overline{h}''$}
Initialize bins $b_1, \dots, b_{K} \leftarrow 0$\;
Initialize counts $c_1, \dots, c_{K} \leftarrow 0$\;
\For{$i \gets 1$ \KwTo $N$}{
 $k \gets \bigl\lfloor y_i \times K \bigr\rfloor$\;  
 $b_k \gets b_k + h_i$\;
 $c_k \gets c_k + 1$\;
}
\For{$k \gets 1$ \KwTo $K$}{
 $\overline{h}_k \gets b_k / c_k$\;
}
Apply Gaussian smoothing to the sequence $\{\overline{h}_k\}_{k=0}^K$\;
$\overline{h}''=(\overline{h}(k)-\overline{h}(k-1))-(\overline{h}(k-1)-\overline{h}(k-2))$\;
\Return{$\overline{h}''$}
\caption{Compute Second Derivative of the Averaged Health Index}
\label{al:aHI}
\end{algorithm}

As shown in Fig.~\ref{fig:2nd_dev_health_idx}, we observe varying degradation rates across the lifecycle: initial slow decline transitions to accelerated deterioration before toward failure. Based on this pattern, we empirically define three lifecycle intervals:
\begin{itemize}
    \item Sluggish stage (0\%–33\%): The second derivative maintains negative values (close to zero), indicating stable system dynamics. This results in a slow decrease of the HI.

    \item Accelerated stage (33\%–85\%): The second derivative shows persistent decrease (becoming more negative). This drives accelerated downward trends in the first derivative, leading to decelerating decay patterns in the HI curve.

    \item Terminal stage (85\%–100\%): The second derivative begins increasing (less negative) while remaining in negative territory. This causes HI decay to approach linear behavior with highest degradation rate.
\end{itemize}

%Drawing from the derived health index, we empirically determine life-cycle ranges of (0, 33\%), (33\%, 85\%], and (85\%, 100\%) to represent the sluggish, moderate, and accelerated degradation stages, respectively.  

\subsubsection{Identifying Target Degradation Stages}
In the target domain, classifying into the three degradation stages becomes a challenge due to the absence of labels. To address this, we propose an unsupervised clustering approach based on the \( k \)-means algorithm for the target domain data. To consider the temporal sequences in the data, soft dynamic time warping is integrated as a distance metric in the \( k \)-means algorithm. After the clustering process, it can be challenging to ascertain the specific degradation stage (e.g., sluggish, accelerated, or terminal) of each target cluster in the absence of labels. We posit that there is a correlation between degradation speed and the variance of sensor readings. Specifically, a faster degradation speed should correspond to higher variance values, and conversely, slower degradation should link to lower variance values. In our study, we begin by computing the variance for each sensor within its respective cluster and sum up variance of all sensors to get the variance of its respective cluster. This is represented by:

\begin{equation}
VAR_c^m = \frac{1}{N_cL}\sum_{i=1}^{N_cL}(x_i - \mu_c)^2,
\label{eq:var}
\end{equation}
where $N_c$ is the number of samples in a cluster $c$, $L$ is the sequence length and $\mu_c$ is the mean value of all element in the cluster.
Subsequently, we determine the overall variance for each cluster as: 
\begin{equation}
VAR_c = \sum_{m=1}^M VAR_c^m.
\label{eq:sum_var}
\end{equation}

\begin{comment}
\begin{equation}
X_c^m = USV^T, 
\label{eq:svd}
\end{equation}
where \( U \) and \( V \) are orthonormal matrices, and \( S \) is a diagonal matrix containing \( K \) singular values, represented as \( \sigma_i \). We then normalize these singular values to yield \( \overline{\sigma}_i \) using the formula 
\begin{equation}
\overline{\sigma}_i = \frac{\sigma_i}{\sum_k \sigma_i}. 
\end{equation}
Subsequently, we determine the overall entropy for each cluster as: 
\begin{equation}
H_c = \sum_{m=1}^M \overline{\sigma}_i \log \overline{\sigma}_i. 
\label{eq:sum_svd_entropy}
\end{equation}
\end{comment}

Finally, we categorize the clusters based on overall variance values. As such, the cluster with the highest variance corresponds to the terminal stage, while the cluster with the lowest variance indicates the sluggish stage.

Due to the increase in computational cost for this clustering approach, we conduct a theoretical analysis of the computational complexity. The theoretical time complexity of the K-means using soft-DTW as the metric is dominated by two key components. First, the pairwise soft-DTW computation between two  $M$ dimensional time series of length $L$  requires  $O(M \cdot L^2)$ operations, as it involves filling a  $L \times L$ dynamic programming matrix with $M$ dimensional distance calculations at each cell.
During the cluster assignment step, all $N$ samples are compared to each of the $K$ cluster centers across  $I$ iterations. This results in a total complexity of $O(I \cdot N \cdot K \cdot M \cdot L^2)$ for distance calculations. Assuming the center update process has negligible complexity compared to distance computations, the simplified overall time complexity becomes $O(I \cdot N \cdot K \cdot M \cdot L^2)$.

\begin{comment}
As varying degradation speeds correlate to different entropy values, the Singular Value Decomposition (SVD) entropy~\cite{roberts1999temporal} is leveraged to ascertain the entropy for each cluster. Given the target data \( X_c^m \) in cluster \( c \) from sensor \( m \), its mathematical breakdown is represented as:
\begin{equation}
X_c^m = USV^T, 
\label{eq:svd}
\end{equation}
where both \( U \) and \( V \) are orthonormal, and \( S \) is a diagonal matrix with \( K \) singular values, denoted as \( \sigma_i \). These singular values are normalized to obtain \( \overline{\sigma}_i \) by the relation \( \overline{\sigma}_i = \frac{\sigma_i}{\sum_k \sigma_i} \). The subsequent SVD entropy for \( X_c^m \) is derived as:
\begin{equation}
H_c^m = -\sum_{i=1}^K \overline{\sigma}_i \log \overline{\sigma}_i. 
\label{eq:svd_entropy}
\end{equation}

For the computation of the aggregate SVD entropy of target data \( X_c \) across all \( M \) sensors in cluster \( c \), the entropies \( H_c^m \) are summed:
\begin{equation}
H_c = \sum_{m=1}^M H_c^m. 
\label{eq:sum_svd_entropy}
\end{equation}

By analyzing the SVD entropy \( H_c \) values, the degradation stage of clustered target data is inferred. Specifically, an increasing trend in \( H_c \) values sequentially represents the sluggish, moderate, and accelerated degradation stages.
\end{comment}

\subsection{Overall Objective}
In the TACDA algorithm (Algorithm \ref{al:pairing}), the primary objective is to fine-tune the target encoder, \(E_T\), for the RUL estimation of equipment. As indicated in Line 1, the algorithm begins with the pretraining of the source encoder $E_S$ and the predictor $R$ using the source domain data $(X_S, y_S)$. After pretraining, the algorithm starts with a domain adaptation utilizing both source and target domains. Following this, data from both domains are clustered into degradation stages---the source domain based on its inherent labels (Line 3) and the target domain using a combination of K-means and variance (Line 4). A subsequent refinement occurs through a second round of domain adaptation for each degradation stage (Lines 5-6). The culmination of the process is the extraction of a trained \(E_T\), which, when coupled with an RUL predictor \(R\), can proficiently estimate the RUL for the target domain.

% 2nd round DA is for fine-tuning the parameters of target encoder $E_T$, target decoder $D_T$ and discriminator $D$ from 1st round DA. After training the target encoder $E_T$, it can be used for target domain RUL estimation with the RUL predictor $R$, i.e., $\widehat{y} = R(E_T(X_T))$.
\begin{algorithm}[h]
\DontPrintSemicolon
\KwIn{Source domain: $\{X_S, y_S\}$,
      Target domain: $X_T$}
\KwOut{Trained target encoder $E_T$}
$E_S, R \gets pretrain(X_S, y_S)$ using Eq.~(\ref{eq:mse_loss})\;
$E_T, D_T, D \gets TA(X_S, X_T, E_S)$  \tcp{Algorithm 1 for 1st round DA}
$X_{S1}, X_{S2}, X_{S3} \gets$ clustering $X_S$ by degradation stage based on $y_S$\;
$X_{T1}, X_{T2}, X_{T3} \gets$ clustering $X_T$ by degradation stage based on K-means and variance\;
\For{$c$ in number of stages}{
  $E_T, D_T, D \gets TA(X_{Sc}, X_{Tc}, E_S)$ \tcp{2nd round DA}
}
\Return{$E_T$}
\caption{Our Proposed TACDA}
\label{al:pairing}
\end{algorithm}

\section{Experiments}
\label{experiments}

\subsection{Preparation of Data}
We employ the C-MAPSS benchmark dataset to evaluate the performance of our proposed method. The C-MAPSS dataset includes operational data from four different turbofan engines. As presented in Table~\ref{tab:cmapss}, each engine has a unique set of operational conditions and fault modes, consisting of 21 sensors deployed in multiple locations to monitor the health of an engine. We pre-process the datasets following the data preparation~\cite{ragab2020contrastive}. The pre-processed data has selected 14 sensors and label is remaining useful life circle of engines.

% shown in Figure~\ref{fig:cmpass}

\begin{table}[]
\centering
\caption{Details of C-MAPSS dataset.}
\label{tab:cmapss}
\begin{tabular}{l|cccc}
\hline
Sub-dataset                    & FD001 & FD002 & FD003 & FD004 \\ \hline
\# Engine units for training        & 100   & 260   & 100   & 249   \\ \hline
\# Engine units for testing       & 100   & 259   & 100   & 248   \\ \hline
%\# Training samples        & 17731 & 48558 & 21220 & 56815 \\ \hline
%\# Testing samples         & 100   & 259   & 100   & 248   \\ \hline
%\# Max life spans (cycles) & 362   & 378   & 512   & 128   \\ \hline
\# Operating conditions    & 1     & 6     & 1     & 6     \\ \hline
\# Fault modes             & 1     & 1     & 2     & 2     \\ \hline
\end{tabular}
\end{table}

% \begin{figure}[t]
% \centering
%  \includegraphics[width=0.4\textwidth]{images/engine.pdf}
% \caption{Diagram of the engines in C-MAPSS data.}
% \label{fig:cmpass}
% \end{figure}

\subsection{Experimental Setting}
All experiments run five times and the average results are shown to prevent the effect of random initialization. Additionally, we set batch size as 256, optimizer as Adam, learning rate as 5e-5 for the target encoder and the domain discriminator. Since the decoder is trained from scratch during the adaptation process, we opt a higher learning rate of 5e-3 specifically for the decoder. And we set the smoothing parameter $\gamma$ for soft-DTW as 0.1. Furthermore, we built and trained our model based on Pytorch and NVIDIA GeForce RTX A4000 GPU. We adopt root mean square error (RMSE) and Score \cite{ragab2020contrastive}. The lower the two indicators are, the better the model is.

The RMSE metric is defined as follows:
\begin{equation}
RMSE = \sqrt{\frac{1}{N}\sum_{i=1}^N (y_i-\widehat{y_i})^2},
\label{eq:rmse}
\end{equation}
where $\widehat{y_i}$ and $y_i$ represent the estimated RUL and true RUL respectively.

The RMSE metric assigns equal importance to both early and late RUL predictions. However, in prognostics applications, late RUL predictions have more detrimental consequences for the systems. In order to address this concern, the Score metric is employed, which imposes a more severe penalty for late RUL predictions. The Score metric is expressed as follows:
%The RMSE metric treats the early and late RUL predictions equally. For prognostics applications, late RUL prediction can be more harmful to the systems. To handle this issue, the Score metric is used to impose bitter penalty for late RUL predictions. It can be formalized as follows:
\begin{equation}
Score_i =\left\{\begin{aligned}
 e^{-\frac{\widehat{y_i}-y_i}{13}}-1; \widehat{y_i} < y_i, \\
 e^{\frac{\widehat{y_i}-y_i}{10}}-1; \widehat{y_i} > y_i, 
\end{aligned}\right.
\label{eq:score_i}
\end{equation}

\begin{equation}
Score = \sum_{i=1}^N Score_i.
\label{eq:score}
\end{equation}

\subsection{Comparison with State-of-the-Art Methods}
\begin{table*}[]
\caption{Comparison of the proposed TACDA against benchmark approaches (Up: RMSE; Down: Score). Note that F1 is short for FD001, and F1$\rightarrow$F2 refers to the scenario where FD001 is the source domain and FD002 is the target domain. Bold indicates the best result, and underline indicates the second-best result.}
\label{tab:performance}
\centering
\setlength{\tabcolsep}{4pt}
\begin{NiceTabular}{l|c|c|c|c|c|c|c|c|c|c|c|c|c}
\toprule
Methods & F1$\rightarrow$F2             & F1$\rightarrow$F3             & F1$\rightarrow$F4             & F2$\rightarrow$F1             & F2$\rightarrow$F3             & F2$\rightarrow$F4             & F3$\rightarrow$F1             & F3$\rightarrow$F2             & F3$\rightarrow$F4             & F4$\rightarrow$F1             & F4$\rightarrow$F2             & F4$\rightarrow$F3             & Avg.           \\ \midrule
Source  & 20.62          & 54.12          & 36.80          & 15.00          & 34.91          & 37.65          & 38.31          & 46.10          & 31.44          & 37.66          & 32.98          & 19.25          & 33.74          \\
Coral   & 20.29          & 38.53          & 34.04          & 13.55          & 32.91          & 33.53          & 34.24          & 36.99          & 30.68          & 38.68          & 31.94          & 16.37          & 30.15          \\
DDC     & 20.08          & 39.26          & 34.14          & 13.53          & 33.40          & 33.47          & 30.31          & 30.60          & 24.19          & 37.08          & 31.78          & 16.06          & 28.66          \\
ADARUL    & 20.26          & \textbf{21.73}          & 31.16          & \ul{13.31}          & \ul{28.98}          & 33.08          & 24.33          & 24.02          & \ul{21.68}          & 26.41          & 25.66          & \ul{15.02} & \ul{23.80}  \\
%ADARUL    & 20.64          & \ul{38.43}          & \ul{31.01}          & 13.65          & \ul{33.51}          & 34.06          & 24.80          & 24.96          & \ul{22.00}          & 30.18          & 29.69          & \ul{14.41}          & 26.45          \\
CLUDA   & 44.36          & 39.17          & 44.71          & 40.07          & 39.17          & 44.71          & 40.09          & 44.45          & 44.72          & 40.14          & 44.38          & 39.19          & 42.10          \\
SDAT    & 20.03          & 39.18          & 34.85          & 17.94          & 35.92          & \ul{31.74}          & 23.90          & 32.11          & 23.16          & \ul{19.81}          & 20.53          & 17.22          & 26.37          \\
CADA    & 20.42          & 38.61          & 31.29          & 13.44          & 33.58          & 35.26          & \ul{20.39}          & \ul{20.85}          & 22.03          & 20.26          & \textbf{18.62} & \textbf{14.34} & 24.09          \\ 
WIDAN & 20.90 & 37.78 & 30.29          & 14.78 & 35.91 & 30.05 & 21.10 & 32.89 & 22.79 & 25.54 & 31.78         & 16.78        & 26.66 \\
DARE-GRAM  &  \textbf{15.70} & 36.32 & \textbf{30.03}         & 17.44 & 30.46 & 31.47 & 21.23 & 22.19 & 24.56 & 20.95 & 24.50         & 15.97        & 24.24   \\ \midrule
TACDA   & \ul{19.39} & \ul{23.16} & \ul{30.38} & \textbf{13.11} & \textbf{21.86} & \textbf{26.82} & \textbf{18.90} & \textbf{19.72} & \textbf{20.46} & \textbf{19.06} & \ul{19.54}          & 15.51          & \textbf{20.66} \\ \midrule
Source  & 5454           & 26074          & 20782          & 536      & 4892    & 19796     & 4991    & 72309          & 40770    & 99469    & 62301   & 2461   & 29986    \\
Coral   & 3312           & 8887    & 14481  & \ul{315}   & 4762  & 12986    & 17364   & 105354    & 33777    & 114609  & 63330  & 1246       & 31702  \\
DDC     & 4062           & 12966          & 16324          & 351     & 5156           & 14969          & 20088          & 58604    & 31221   & 90087    & 60334    & 1323     & 26290     \\
ADARUL  &  3624           & \ul{1753}           & 11788          & 330            & 3815           & 16077   & 6250           & 31462          & 7062      & 14645     & 35174          & 754            & 11061          \\
CLUDA   & 22351          & 5869           & 18270          & 6000    & 5963           & 18655          & 5797           & 30091   & 20506    & 6602      & 29771     & 7100    & 14748     \\
SDAT    & \ul{2343}           & 5490          & 19432          & 762            & 10000          & 10587   & 2460     & 27984          & 4399     & \ul{1241}     & \textbf{2793}  & 844     & 7361           \\
CADA    & 4156           & 10540          & 12050          & 321   & 5484           & 19766   & \ul{1856}   & 16760   & 4134  & 1755      & 4949   & 913   & 6890           \\ 
WIDAN & 2787  & 6388  & 10121 & 353   & 4863  & \ul{6983}  & 5733   & 7591  & 6540  & 2950  & 7142  & \ul{554}   & 5167 \\
DARE-GRAM  & 4438  & 8765  & \textbf{8267} & 740   & \ul{4057}  & 9819  & 2139  & \ul{7053}  & \ul{3675}  & 1569  & \ul{2944} & \textbf{566} & \ul{4503} \\ \midrule
TACDA   & \textbf{2183}  & \textbf{1678}  & \ul{10563} & \textbf{298}            & \textbf{1901}  & \textbf{5948}  & \textbf{834}   & \textbf{4770}  & \textbf{2422}  & \textbf{966}   & 4564   & 799   & \textbf{3077}  \\ \bottomrule
\end{NiceTabular}
\end{table*}
We compare proposed TACDA with SOTA UDA methods, including DDC~\cite{tzeng2014deep}, Coral~\cite{sun2017correlation}, ADARUL~\cite{ragab2020adversarial}, CLUDA~\cite{vayyat2022cluda}, SDAT~\cite{rangwani2022closer}, CADA~\cite{ragab2020contrastive}, DARE-GRAM~\cite{nejjar2023dare} and WIDAN~\cite{he2024interpretable}. For fair comparisons, we adopt the same feature extractor (5 layers BiLSTM and 32 hidden dimensions) and predictor for these methods, each of which runs five times to show average results. Further, the results of source only (Source) are also compared.

Table~\ref{tab:performance} shows the RMSE and Score results respectively in 12 cross-domain scenarios for RUL prediction. From the results, we observe that TACDA achieves the best averaged performance across all scenarios with regards to both RMSE and Score, whose improvement is more than 10\% and 32\% for RMSE and Score compared with the second best method. Specifically, TACDA outperforms all the SOTA methods in the 7 scenarios for RMSE and in the 9 scenarios for Score. For example, TACDA outperforms the second-best method by 55\% in FD003$\rightarrow$FD001, by 39\% in FD002$\rightarrow$FD004, and by 32\% in the FD003$\rightarrow$FD002. Furthermore, TACDA attains the second-best performance in 4 scenarios with regards to RMSE and 1 scenario Score. Our proposed TACDA achieves the superior performance with the ability of preserving target domain specific information.

\begin{table*}[]
\caption{Ablation study for the proposed TACDA (Up: RMSE; Down: Score). Note that $C$ denotes the clustering and pairing strategy for consistent degradation alignment, while $D_T$ is the target decoder.}
\label{tab:ablation}
\centering
\setlength{\tabcolsep}{5pt}
\begin{NiceTabular}{l|c|c|c|c|c|c|c|c|c|c|c|c|c}
\toprule
Methods    & F1$\rightarrow$F2             & F1$\rightarrow$F3             & F1$\rightarrow$F4             & F2$\rightarrow$F1             & F2$\rightarrow$F3             & F2$\rightarrow$F4             & F3$\rightarrow$F1             & F3$\rightarrow$F2             & F3$\rightarrow$F4             & F4$\rightarrow$F1             & F4$\rightarrow$F2             & F4$\rightarrow$F3             & Avg.           \\ \midrule
Source     & 20.62          & 54.12          & 36.80          & 15.00          & 34.91          & 37.65          & 38.31          & 46.10          & 31.44          & 37.66          & 32.98          & 19.25          & 33.74          \\
w/o $C$      & \textbf{19.12} & 23.40 & 31.07          & 13.75          & 27.47          & 31.00          & 21.21          & 20.93          & 22.05          & 21.48          & 20.11          & 16.16          & 22.31          \\
w/o $D_T$   & 20.33          & \textbf{21.03} & 31.12          & 13.25          & 26.32          & \textbf{26.36} & 21.70          & 22.46          & 21.85          & 27.67          & 23.83          & 17.04          & 22.75          \\ 
w/o $C$-$D_{T}$ & 20.26          & 21.73          & 31.16          & 13.31          & 28.98          & 33.08          & 24.33          & 24.02          & 21.68          & 26.41          & 25.66          & \textbf{15.02} & 23.80          \\ 
TACDA      & 19.39          & 23.16          & \textbf{30.38} & \textbf{13.11} & \textbf{21.86} & 26.82 & \textbf{18.90} & \textbf{19.72} & \textbf{20.46} & \textbf{19.06} & \textbf{19.54} & 15.51          & \textbf{20.66} \\ \hline
Source     & 5454           & 26074          & 20782          & 536            & 4892           & 19796          & 4991           & 72309          & 40770          & 99469          & 62301          & 2461           & 29986          \\
w/o $C$      & 2288           & 1786  & 10767 & 366            & 2865           & 10558          & 2275           & 9635           & 6428           & 1952           & 7205           & 1050           & 4765           \\
w/o $D_T$   & 3353           & \textbf{1634}  & 11627          & 301            & 4808           & 6268           & 1876           & 9070           & 3824           & 2798           & 13715          & 1122           & 5033           \\ 
w/o $C$-$D_T$ & 3624           & 1753           & 11788          & 330            & 3815           & 16077          & 6250           & 31462          & 7062           & 14645          & 35174          & \textbf{754}            & 11061          \\ 
TACDA      & \textbf{2183}  & 1678           &  \textbf{10563}          & \textbf{298}            & \textbf{1901}  & \textbf{5948}  & \textbf{834}   & \textbf{4770}  & \textbf{2422}  & \textbf{966}   & \textbf{4564}  & 799   & \textbf{3077}  \\ \bottomrule
\end{NiceTabular}
\end{table*}

\subsection{Ablation Study}
\subsubsection{Component Contribution Study}
To validate the contribution of key components, we perform the ablation study for our proposed TACDA. We derive four variants of TACDA, namely, “Source”, “w/o $C$”, “w/o $D_{T}$” and “w/o $C$-$D_{T}$”. Specifically, the variant “Source” denotes the unadapted version of our model. “w/o $C$” is the variant without using the 2nd round DA (i.e., no fine-tuning with the clustered data) for consistent degradation alignment. The variant “w/o $D_{T}$” refers to our adversarial adaptation without using the target decoder $D_{T}$ for target-specific adaptation. The last variant “w/o $C$-$D_{T}$” uses the adversarial loss only, i.e., it has neither fine-tuning step or target decoder $D_{T}$. It is worth noting that “w/o $C$-$D_{T}$” is the same as ADARUL \cite{ragab2020adversarial}.

Table~\ref{tab:ablation} shows presents the comparative outcomes between TACDA and its variants. Our observation reveals that the “Source” exhibits the poorest performance, implying the significant disparity between the source and target domain. The proposed TACDA suppresses the “w/o $C$”, “w/o $D_{T}$” and “w/o $C$-$D_{T}$” in around 35\%, 39\% and 72\%, regarding the average improvements in terms of Score. These results demonstrate the effectiveness of the decoder, as well as the clustering and pairing strategy for the 2nd round DA for RUL prediction.

\subsubsection{Reconstruction Loss Study}
\begin{figure}[]
    \centering
    \includegraphics[width=0.45\textwidth]{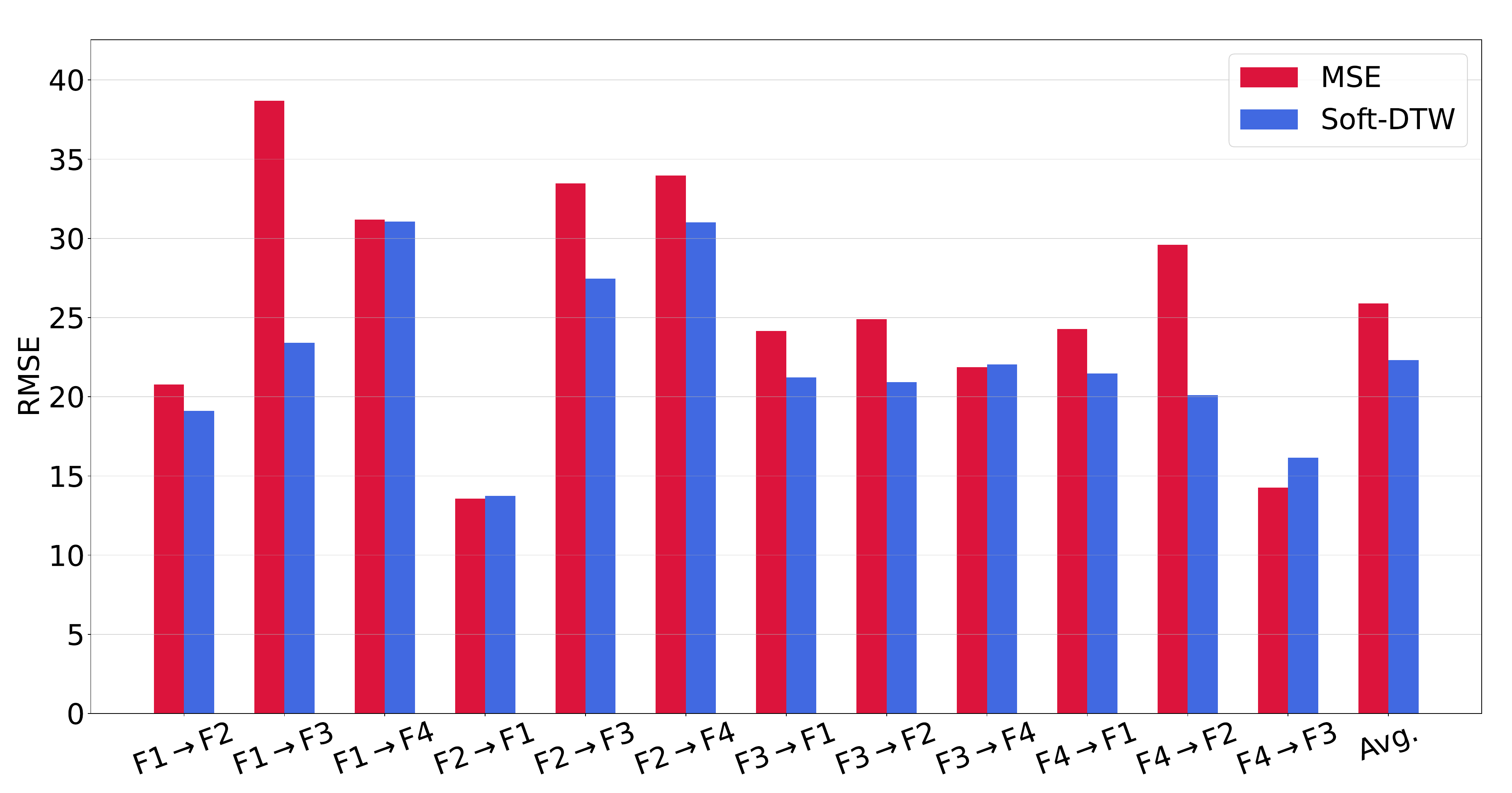}
    \includegraphics[width=0.45\textwidth]{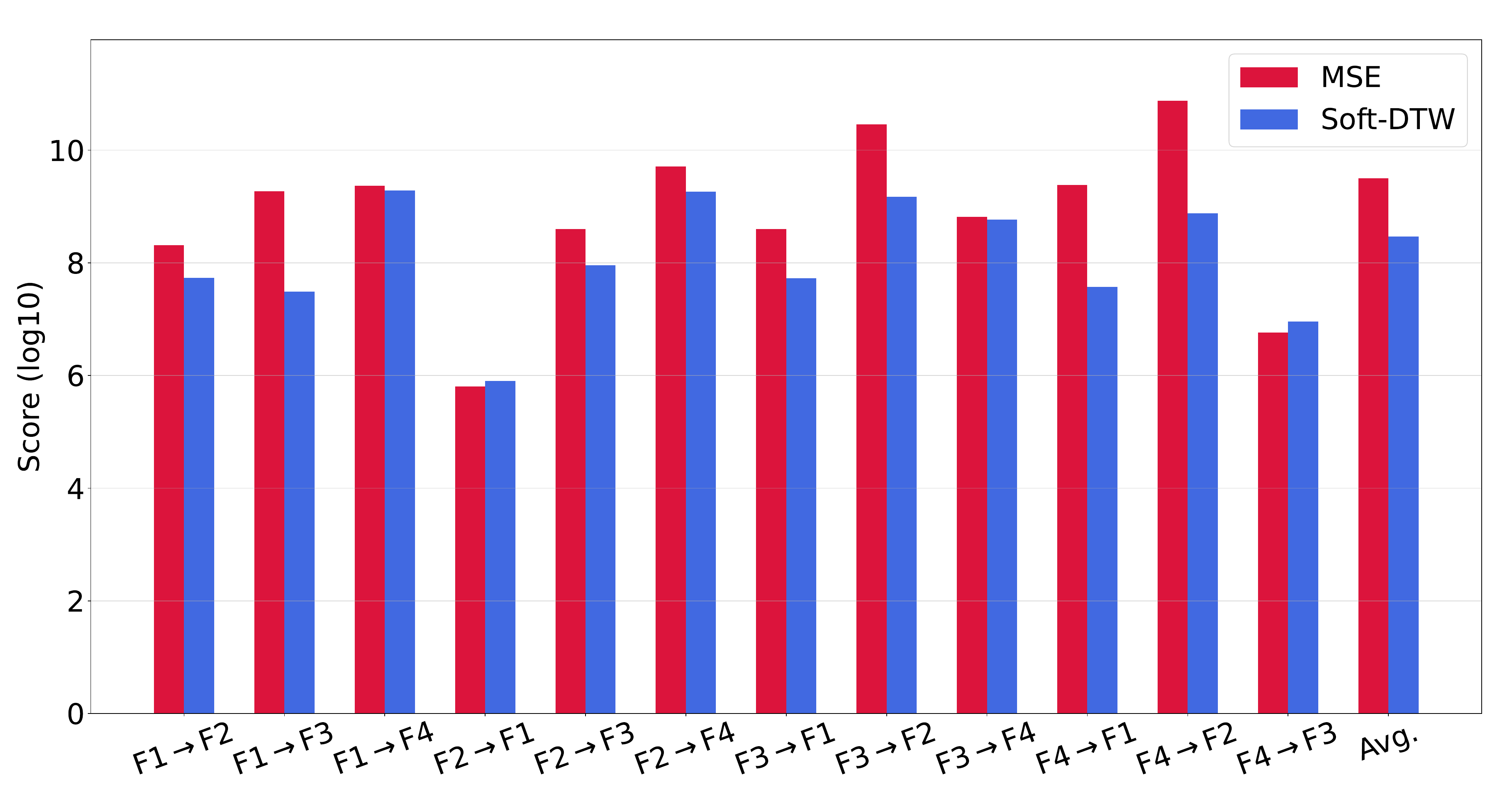}

    \caption{Comparison of using MSE and soft-DTW as reconstruction loss.}
    \label{fig:loss_compare}
\end{figure}

To verify the effectiveness of soft-DTW loss for domain information preservation, we perform comparison between MSE and soft-DTW loss in our proposed TACDA without degradation alignment. As shown in Fig.~\ref{fig:loss_compare}, 
%the performance of soft-DTW loss is around 14\% and 65\% better than MSE loss, in terms of averaged RMSE and Score across all scenarios. Specifically, 
soft-DTW outperforms MSE in 9 scenarios in terms of RMSE and in 10 scenarios in terms of Score. The rationale underlying this phenomenon resides in the capacity of soft-DTW to effectively capture temporal sequence patterns.

\subsection{Soft-DTW Contribution Analysis}
\begin{figure}[]
    \centering
    \includegraphics[width=0.23\textwidth]{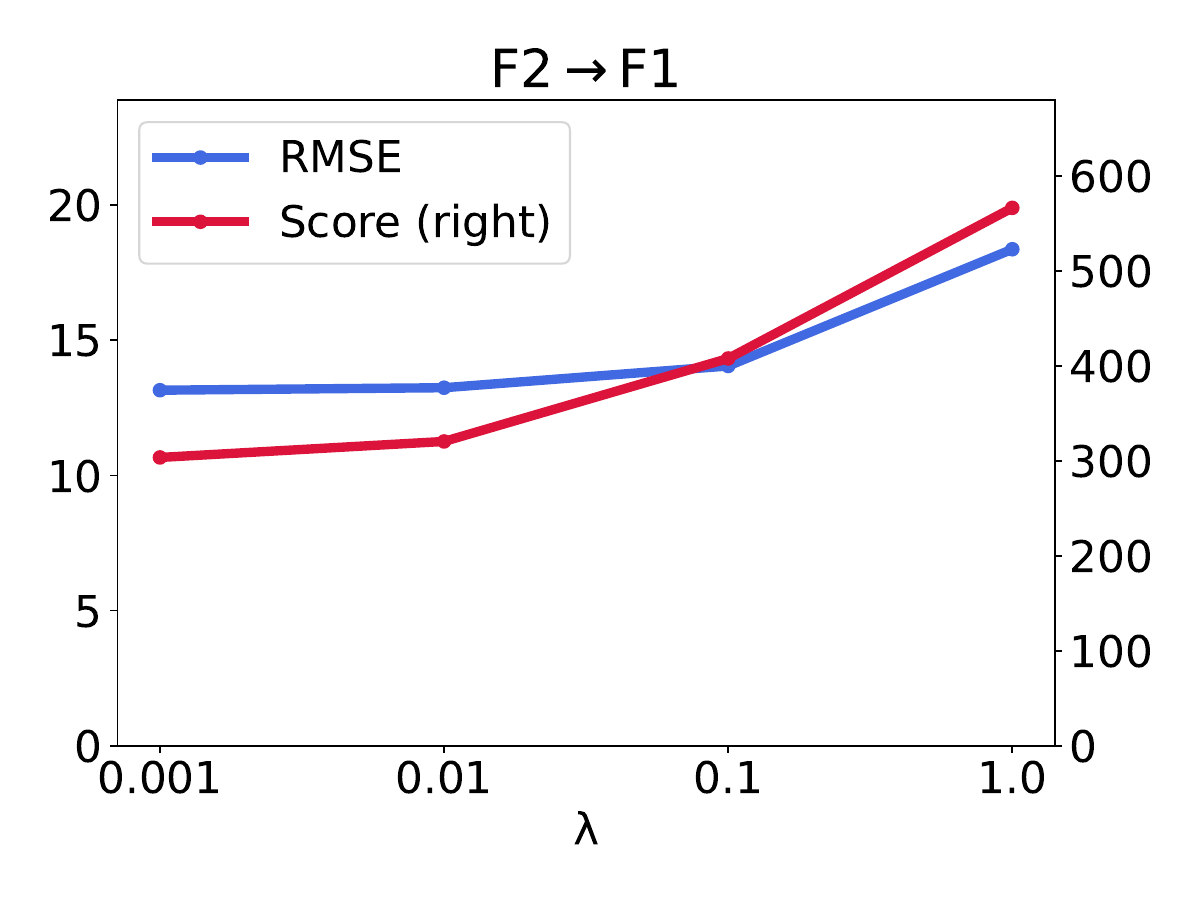}
    \includegraphics[width=0.23\textwidth]{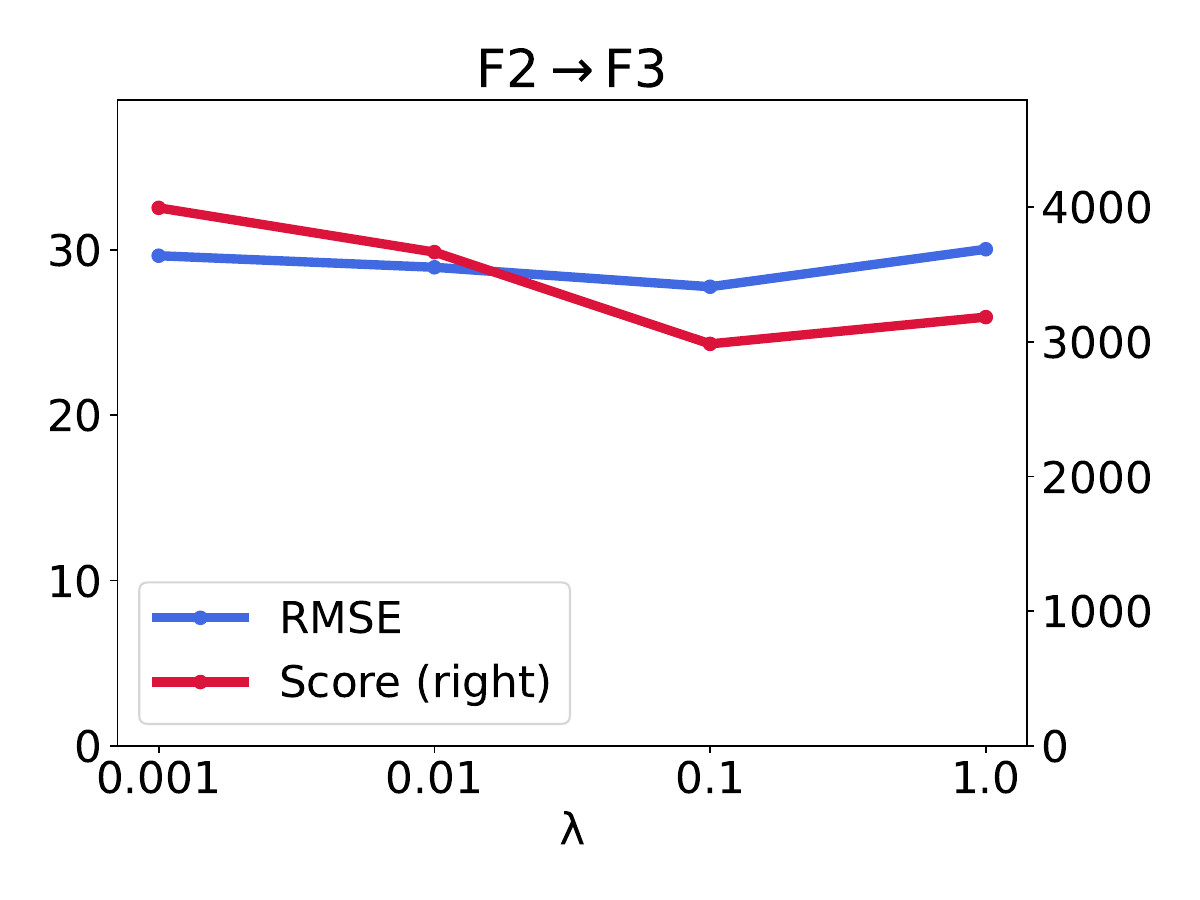}
    \includegraphics[width=0.23\textwidth]{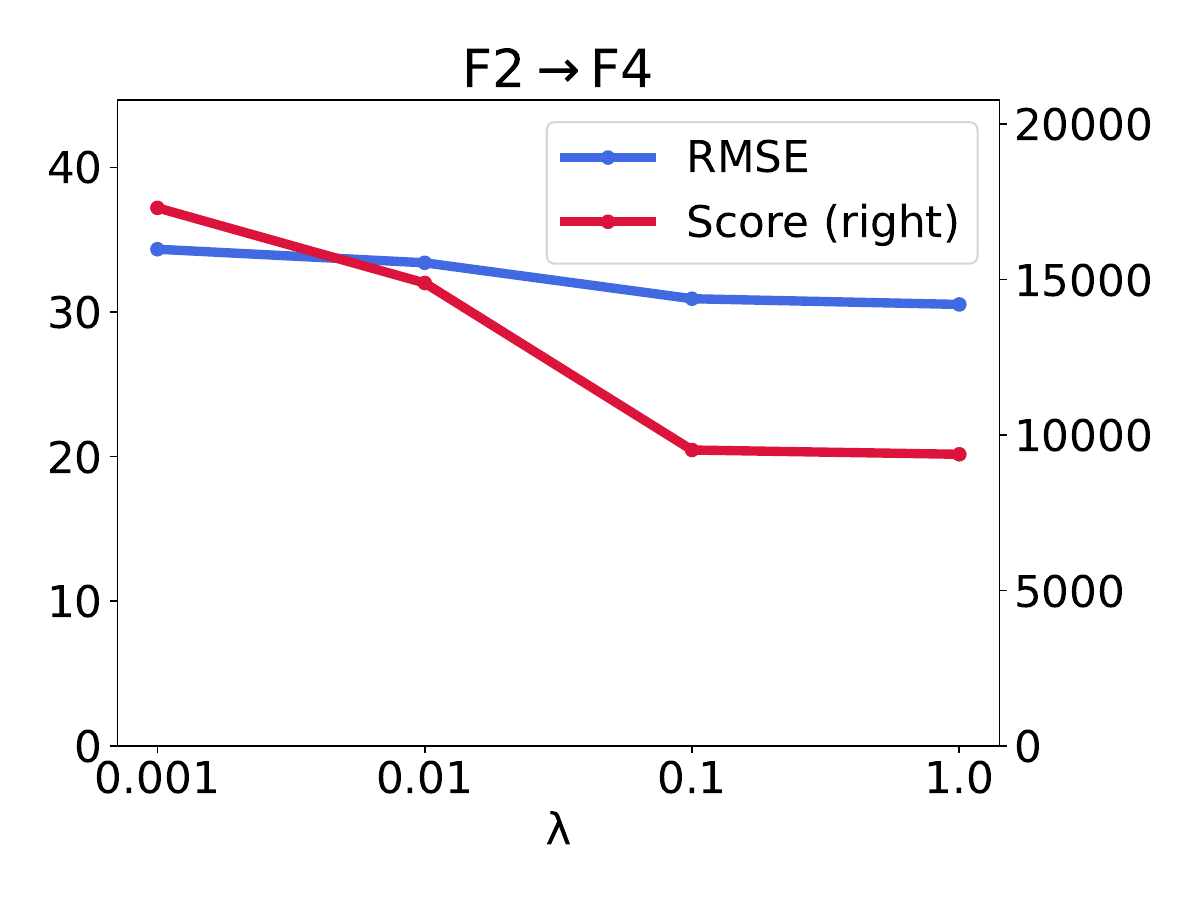}
    \includegraphics[width=0.23\textwidth]{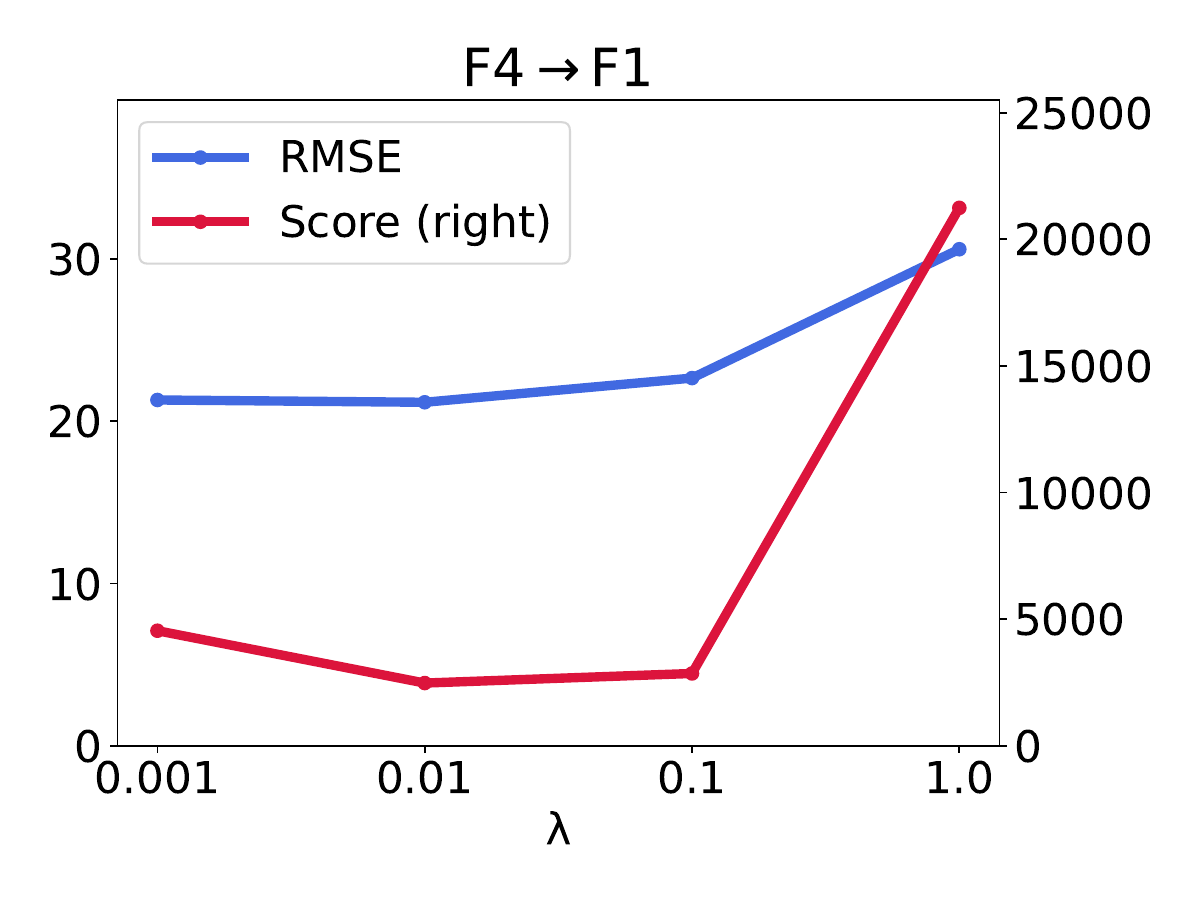}
    \includegraphics[width=0.23\textwidth]{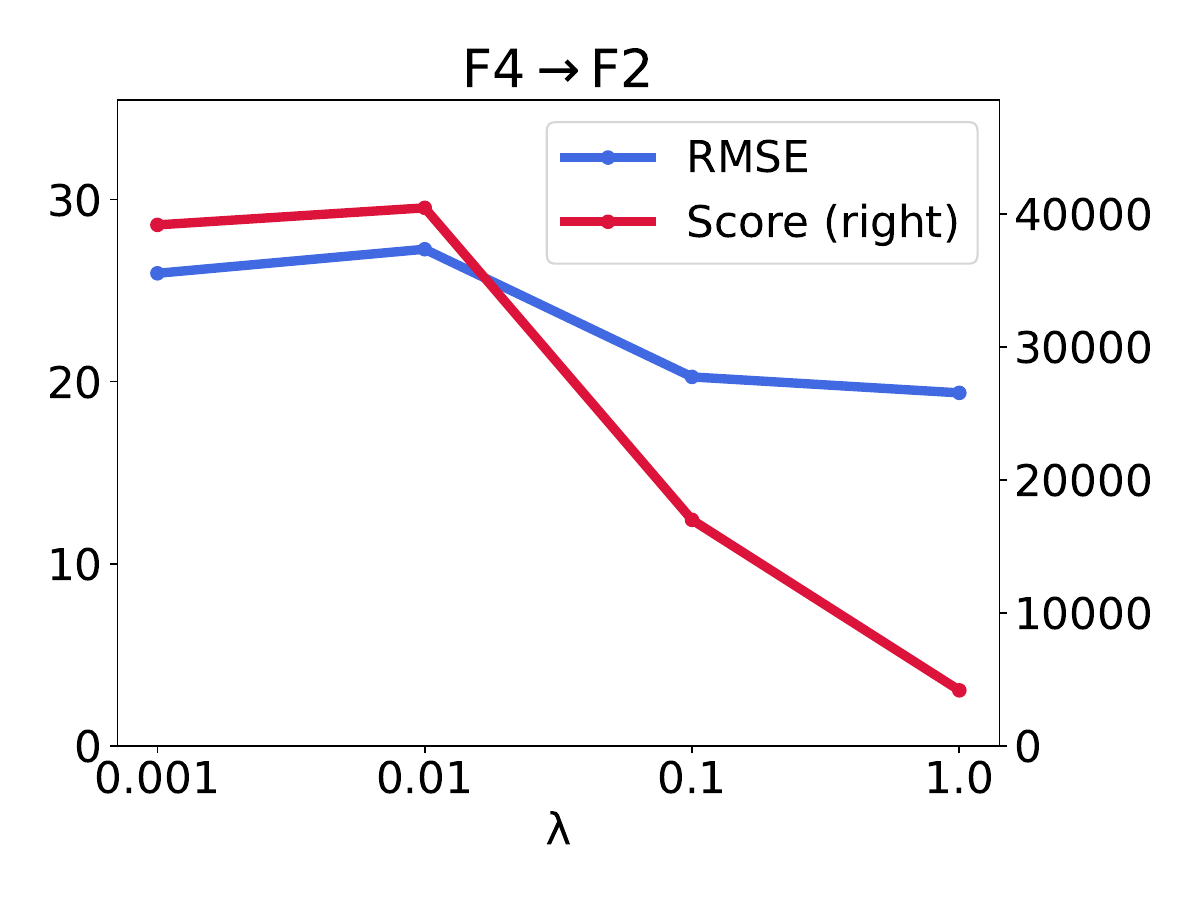}
    \includegraphics[width=0.23\textwidth]{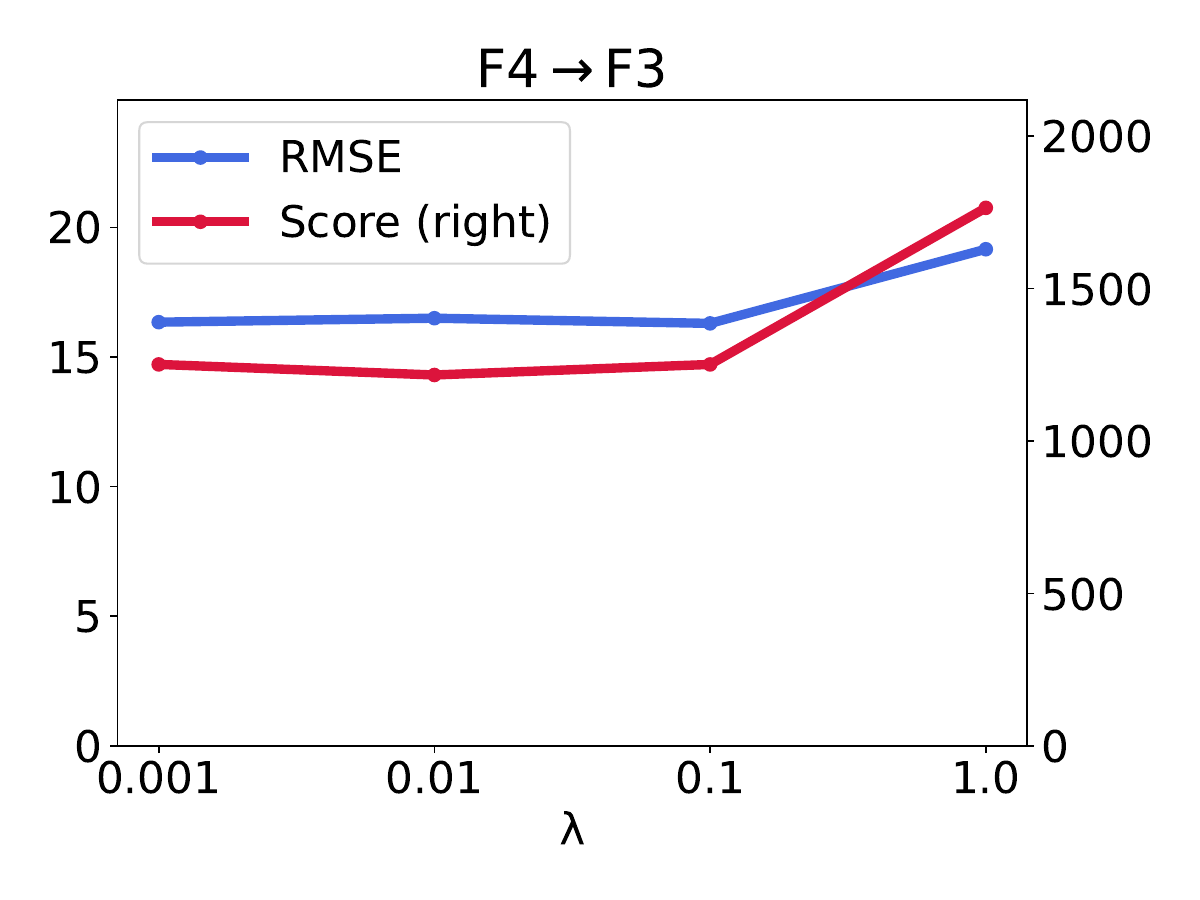}

    \caption{The sensitivity analysis for different $\lambda$ in the 1st round DA.}
    \label{fig:lambda_sensitivity_1stDA}
\end{figure}

\begin{figure}[]
    \centering
    \includegraphics[width=0.23\textwidth]{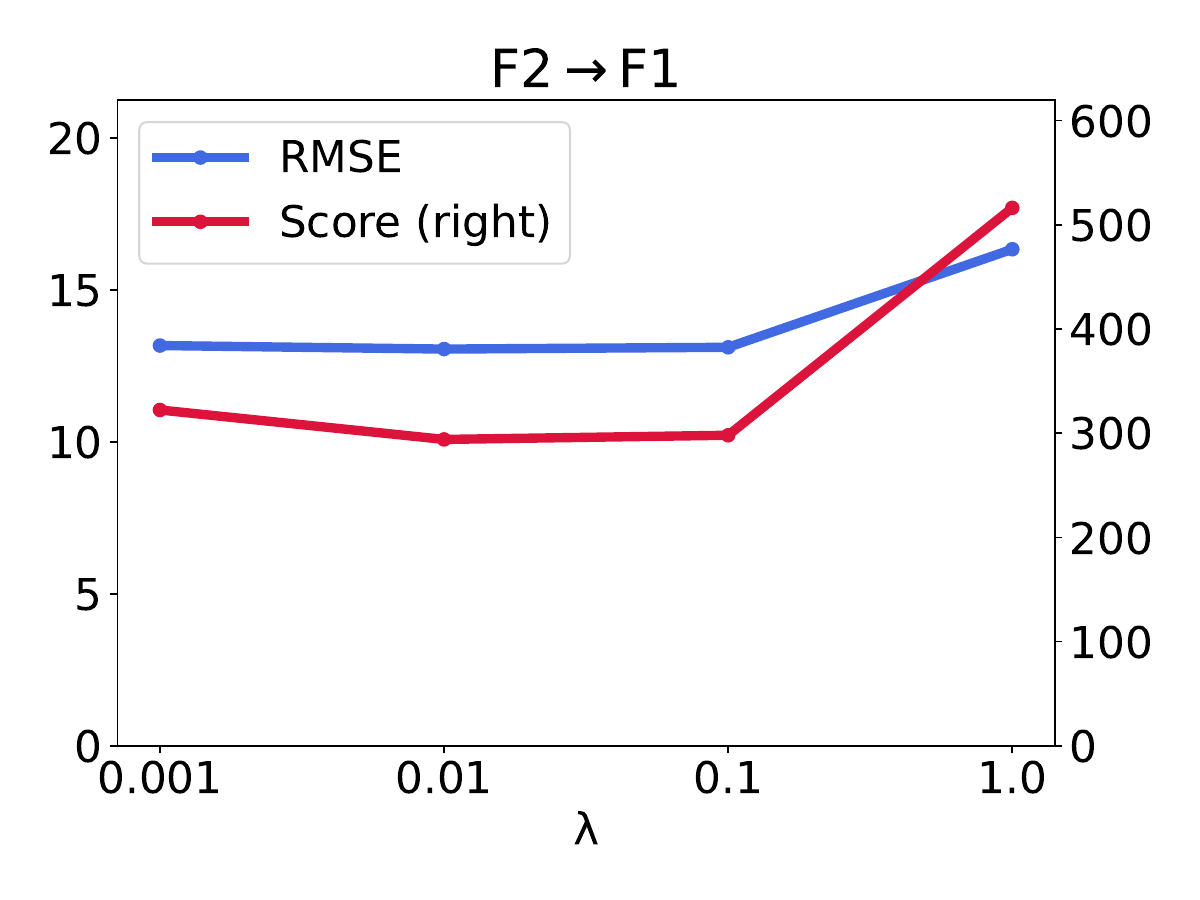}
    \includegraphics[width=0.23\textwidth]{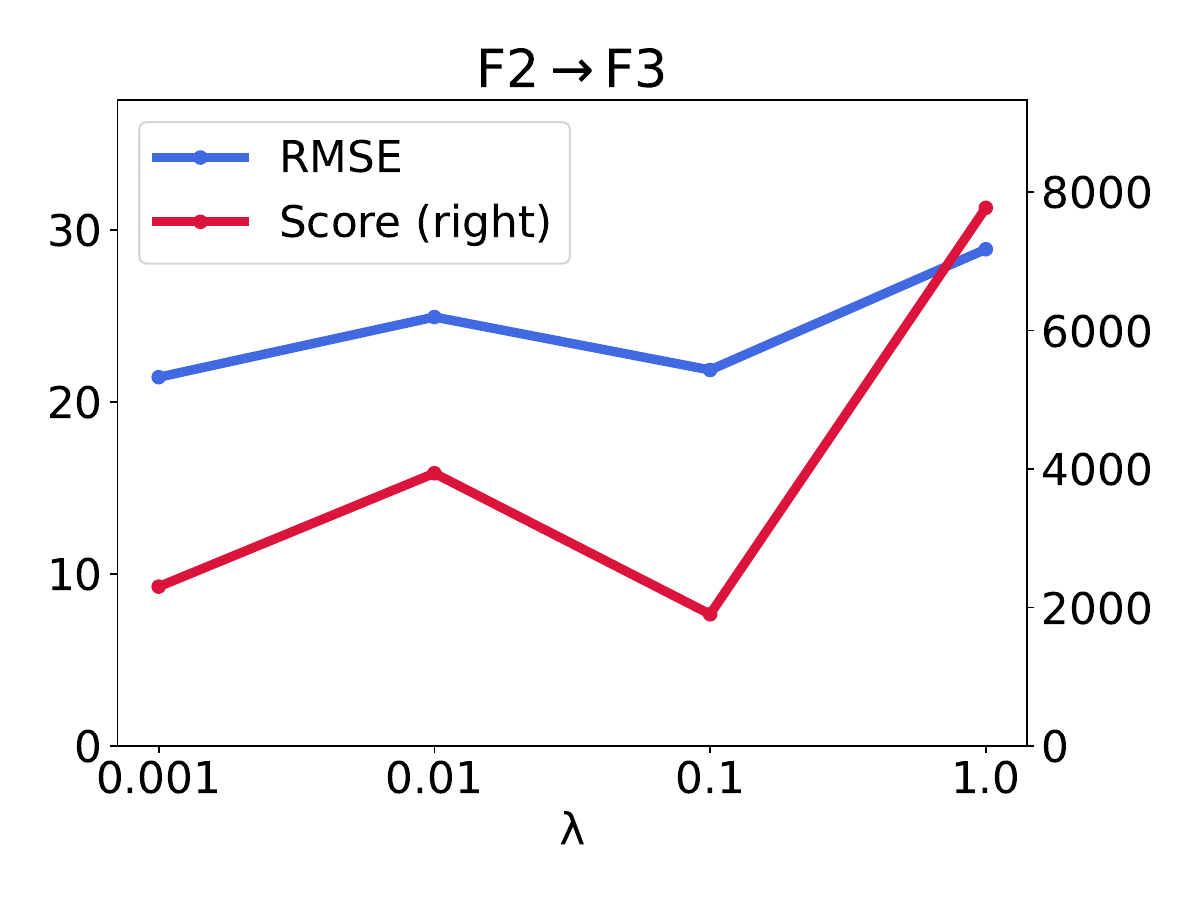}
    \includegraphics[width=0.23\textwidth]{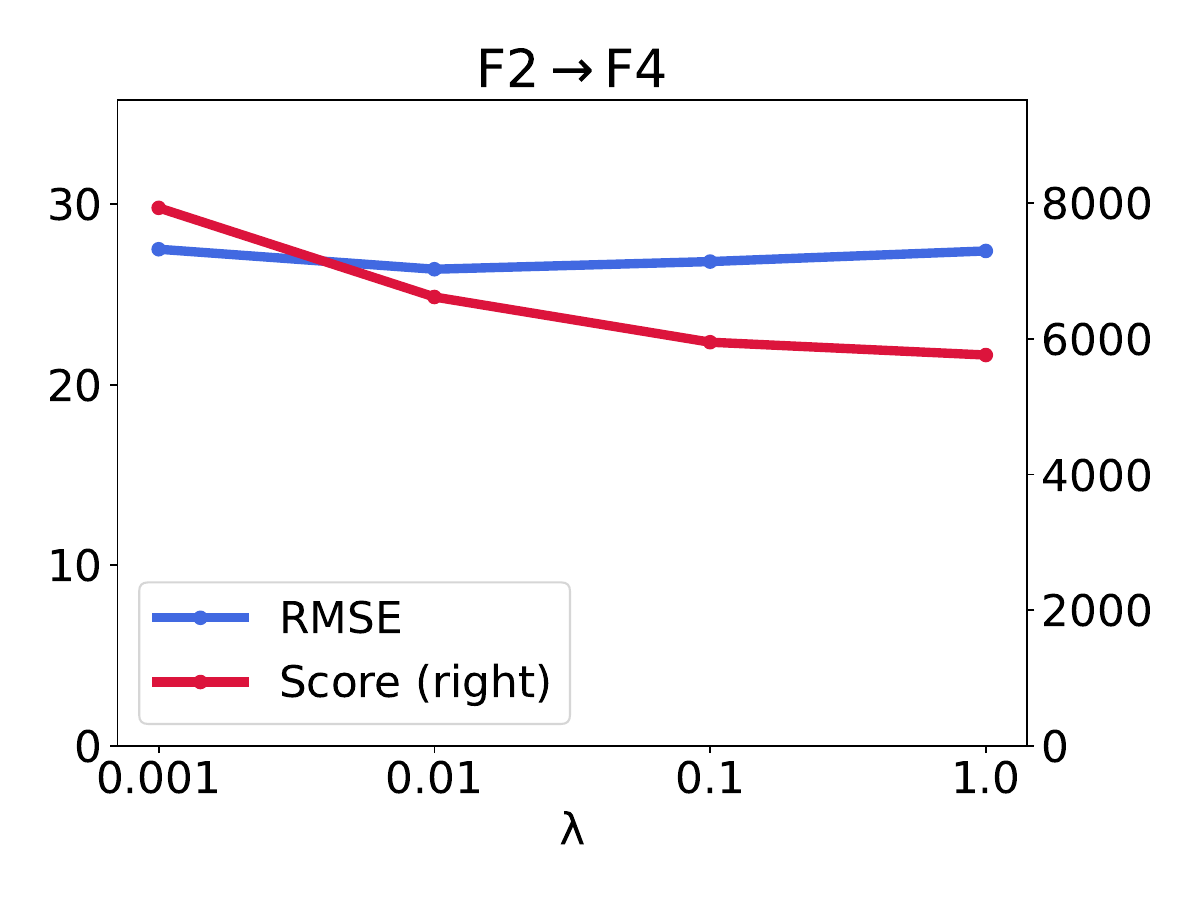}
    \includegraphics[width=0.23\textwidth]{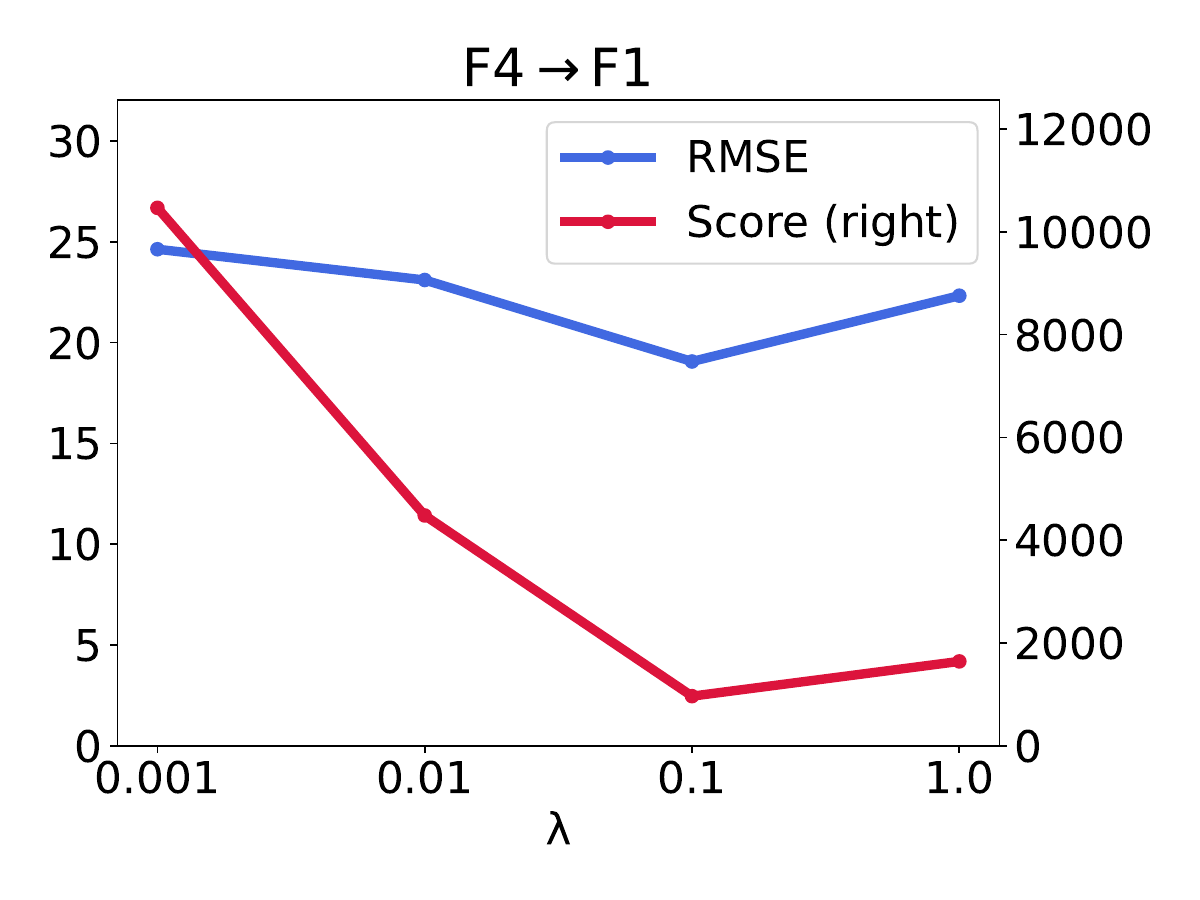}
    \includegraphics[width=0.23\textwidth]{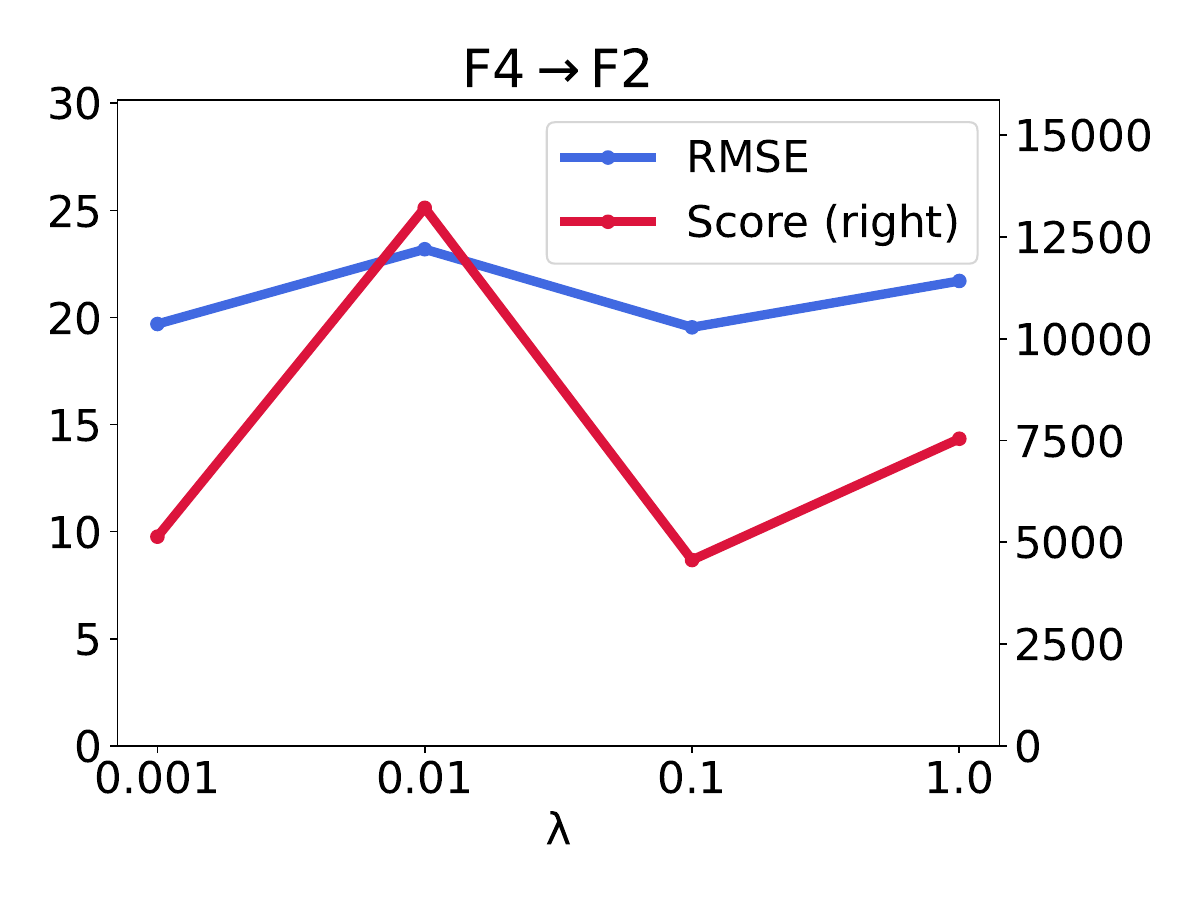}
    \includegraphics[width=0.23\textwidth]{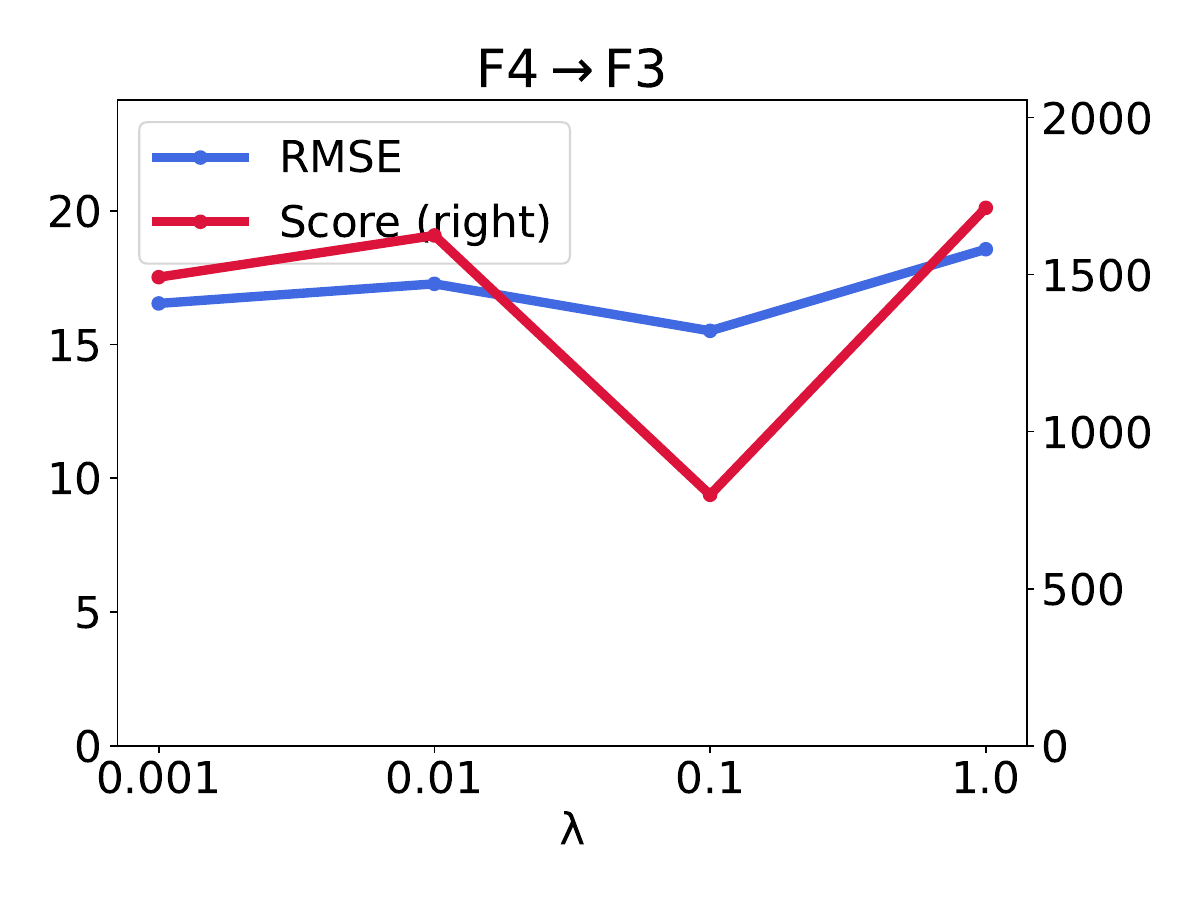}

    \caption{The sensitivity analysis for different $\lambda$ in the 2nd round DA.}
    \label{fig:lambda_sensitivity}
\end{figure}

We conducted a sensitivity analysis of the hyperparameter $\lambda$ to investigate the impact of soft-DTW on the proposed method. Several experiments were carried out using various values of $\lambda$, ranging from 0.001 to 1 with intervals of 10×. Given that the initial magnitude of the soft-DTW surpasses the adversarial loss by a factor of 1000 during the training, it is advisable to constrain the weight parameter $\lambda$ to a value not exceeding 1. Fig.~\ref{fig:lambda_sensitivity_1stDA} and Fig.~\ref{fig:lambda_sensitivity} show the performance of TACDA under different values of $\lambda$ after the 1st and 2nd rounds of DA.
The results indicate that the proposed method demonstrates stability across most cross-domain scenarios in terms of RMSE. However, we observed that the performance of TACDA tends to degrade when $\lambda$ is either too small or too large in terms of Score. 
%For instance, in the FD004 $\rightarrow$ FD001 scenario, TACDA exhibits the worst performance when $\lambda$ is set to 0.001. Conversely, in the FD002 $\rightarrow$ FD003 scenario, TACDA performs worse than other cases when $\lambda$ is set to 1, despite being stable in other instances. 
For instance, in the FD004 $\rightarrow$ FD001 scenario, TACDA exhibits the worst performance when $\lambda$ is set to 1 during the 1st round of DA, and when $\lambda$ is set to 0.001 during the 2nd round. Similarly, in the FD002 $\rightarrow$ FD003 scenario, TACDA performs poorly when $\lambda$ is 0.001 in the 1st round, and when $\lambda$ is 1 in the 2nd round.
To achieve satisfactory performance, we recommend setting the value of $\lambda$ as 0.1. The potential reason for the consistency observed in RMSE while encountering variability in the Score metric could be attributed to the temporal influence exerted on the Score metric during the utilization of soft-DTW temporal alignment.

\begin{figure}[]
    \centering
    \includegraphics[width=0.24\textwidth]{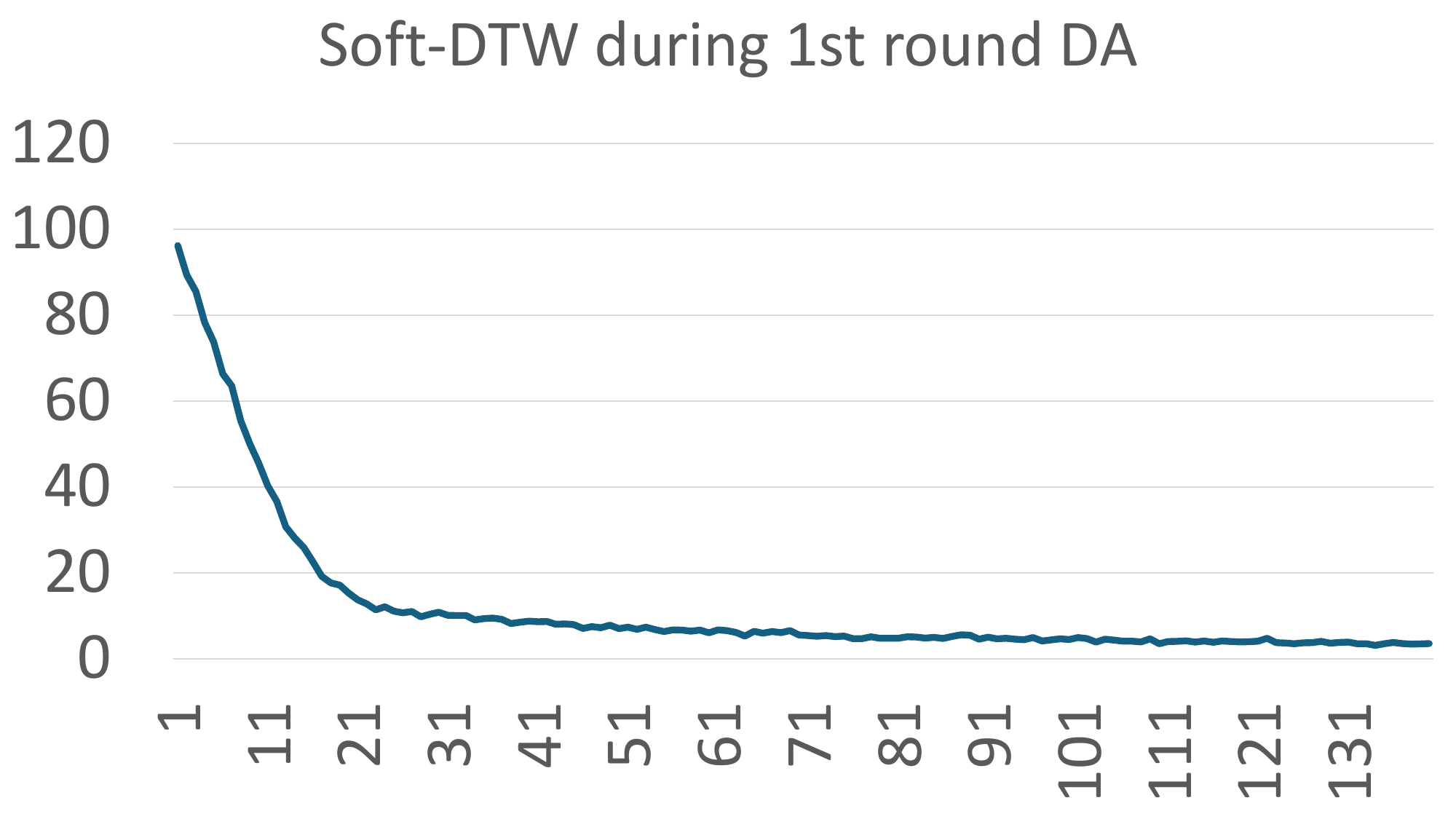}
    \includegraphics[width=0.24\textwidth]{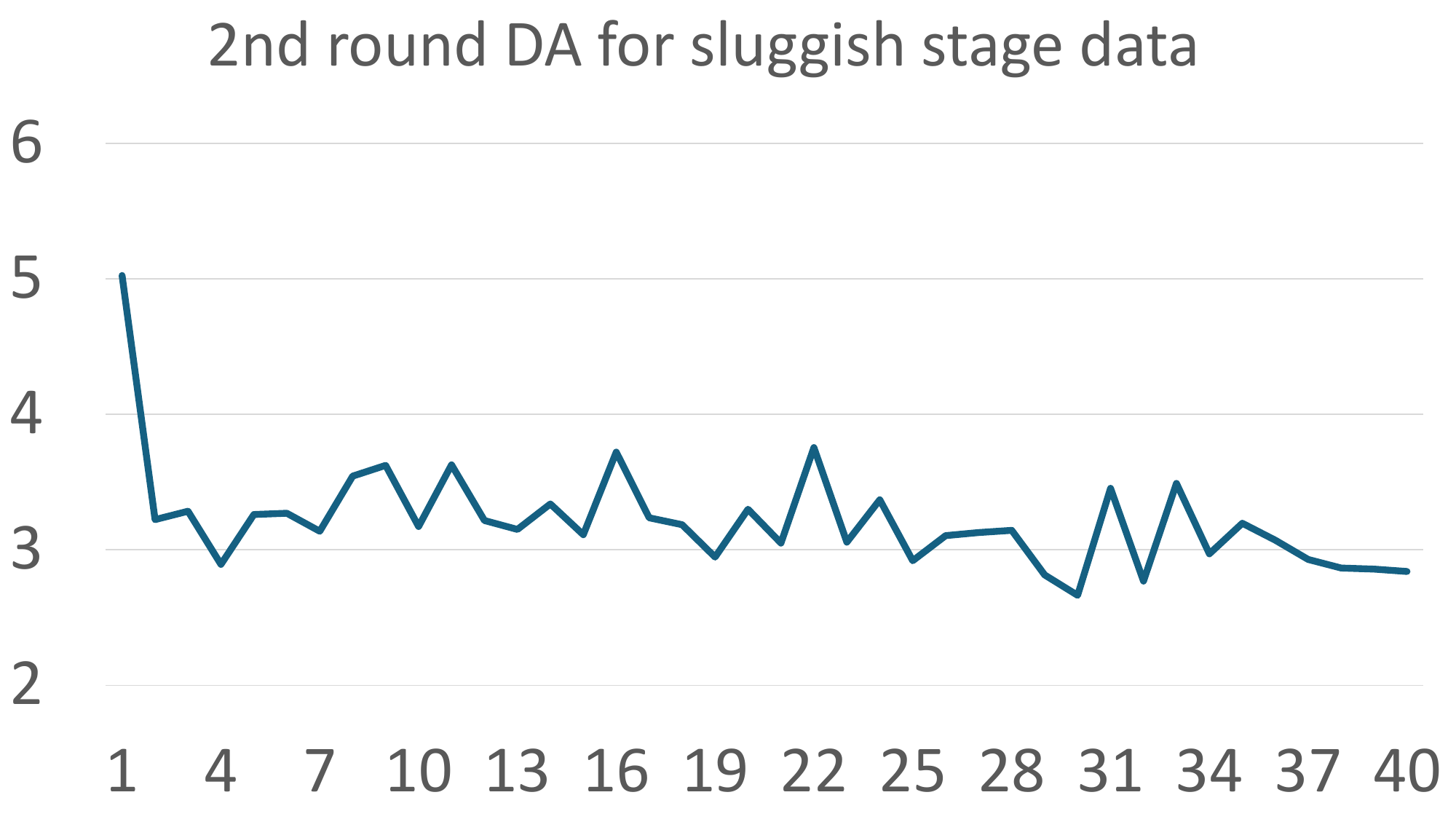} 
    
    \vspace{2mm}
    
    \includegraphics[width=0.24\textwidth]{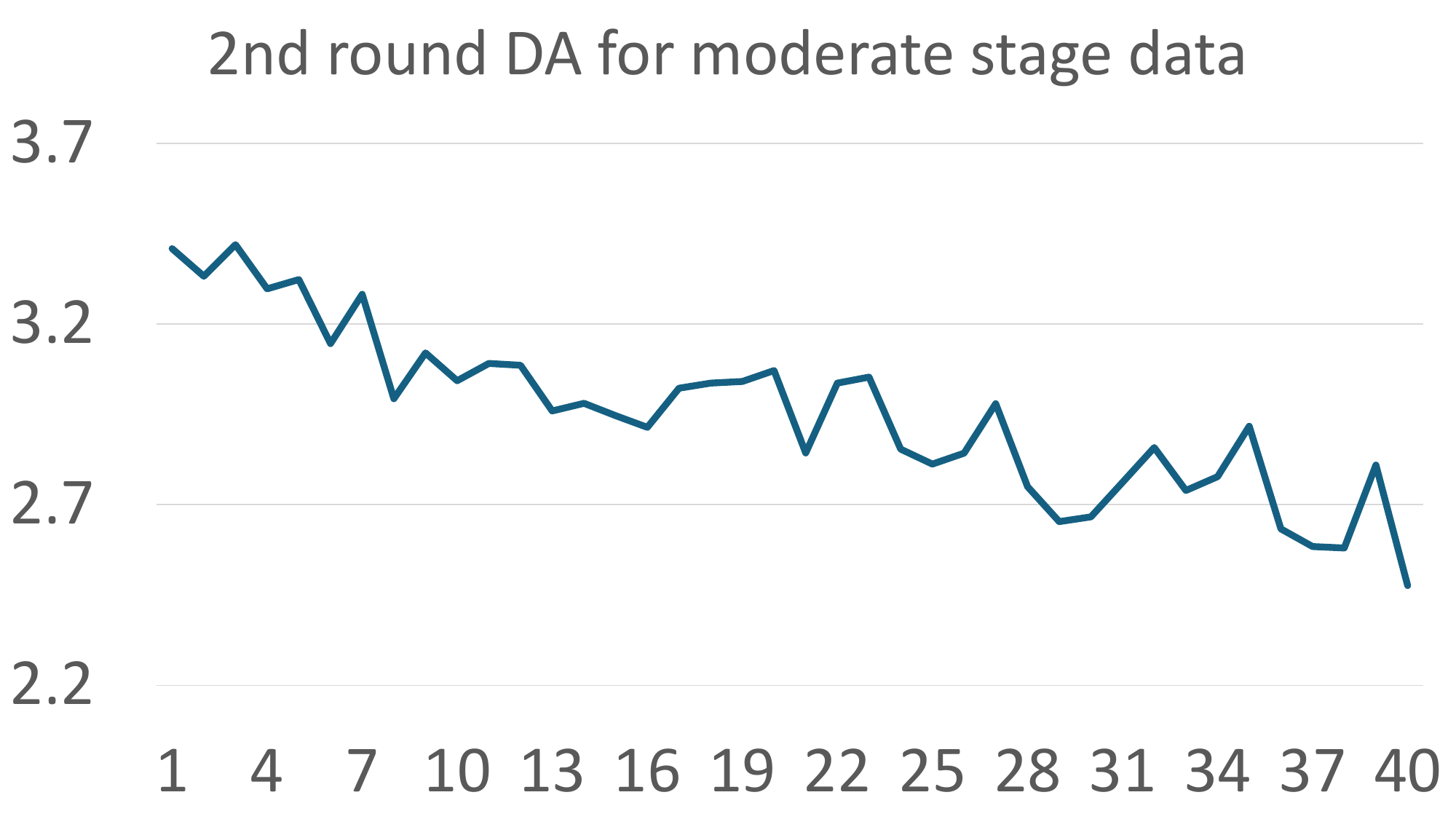}
    \includegraphics[width=0.24\textwidth]{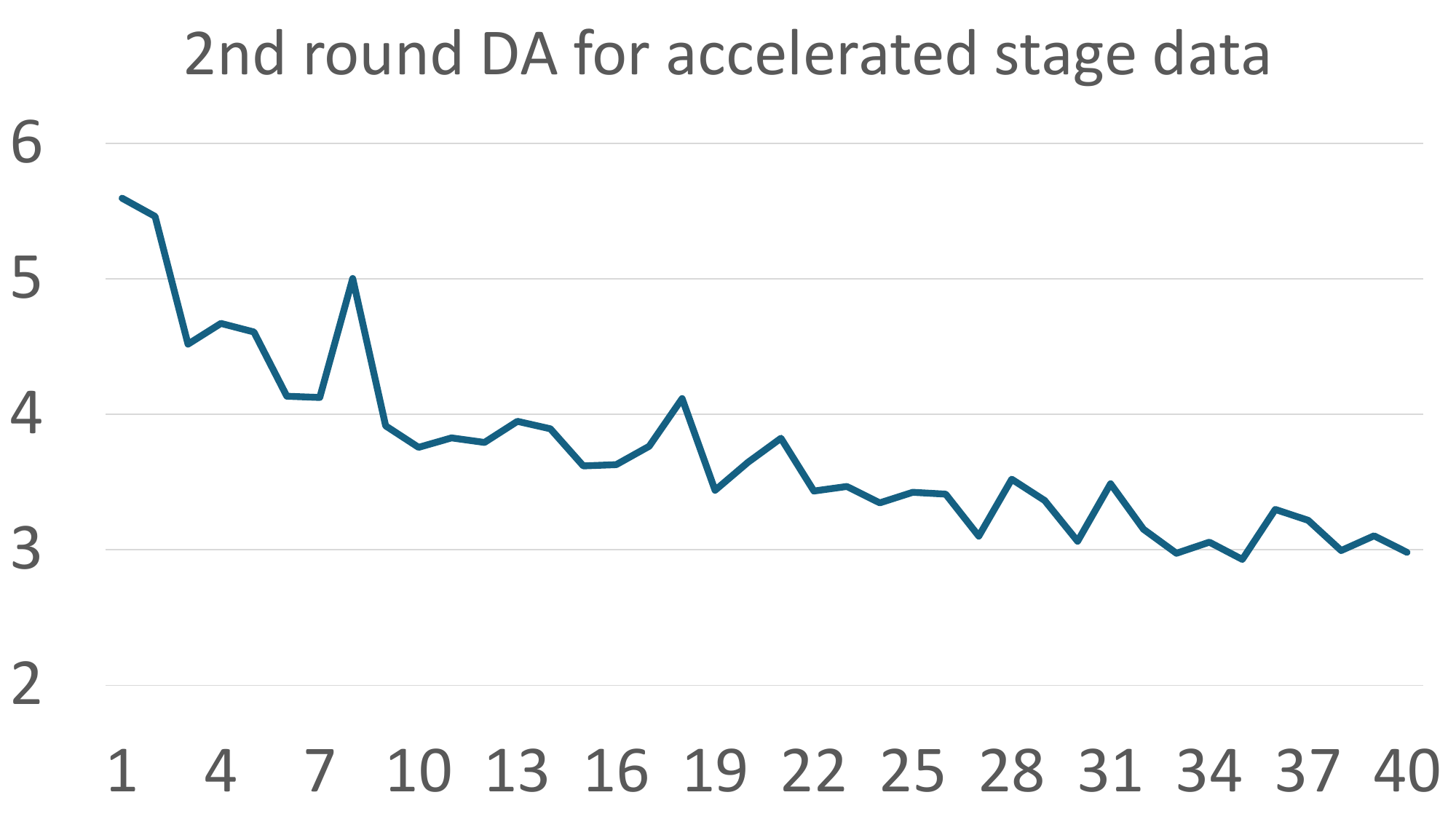}

    \caption{The changes of soft-DTW loss in the 1st and 2nd round DA.}
    \label{fig:soft-DTW_train}
\end{figure}

To highlight the role of soft-DTW as a reconstruction loss in preserving target-specific information, we analyzed its progression over training epochs during both the 1st and 2nd rounds of DA. As illustrated in Fig.~\ref{fig:soft-DTW_train}, the soft-DTW loss exhibits a sharp decline in the initial epochs of the 1st round of DA, followed by a gradual stabilization as training converges. Since soft-DTW serves as a reconstruction loss, this behavior underscores the effectiveness of the proposed approach in maintaining target-specific information while facilitating cross-domain alignment.

In the 2nd round of DA, during the initial stages, the soft-DTW loss values in the sluggish stage and accelerated stage are higher than those in the moderate stage, indicating that the information from the sluggish and accelerated stages was not well preserved during the 1st round of DA. Through the 2nd round of DA, the soft-DTW loss continues to decrease across data from various degradation stages, further demonstrating its ability to capture and preserve distinct target-specific characteristics at different stages.

\subsection{Insufficient Degradation Data Analysis}
\begin{figure}[]
    \centering
    \includegraphics[width=0.45\textwidth]{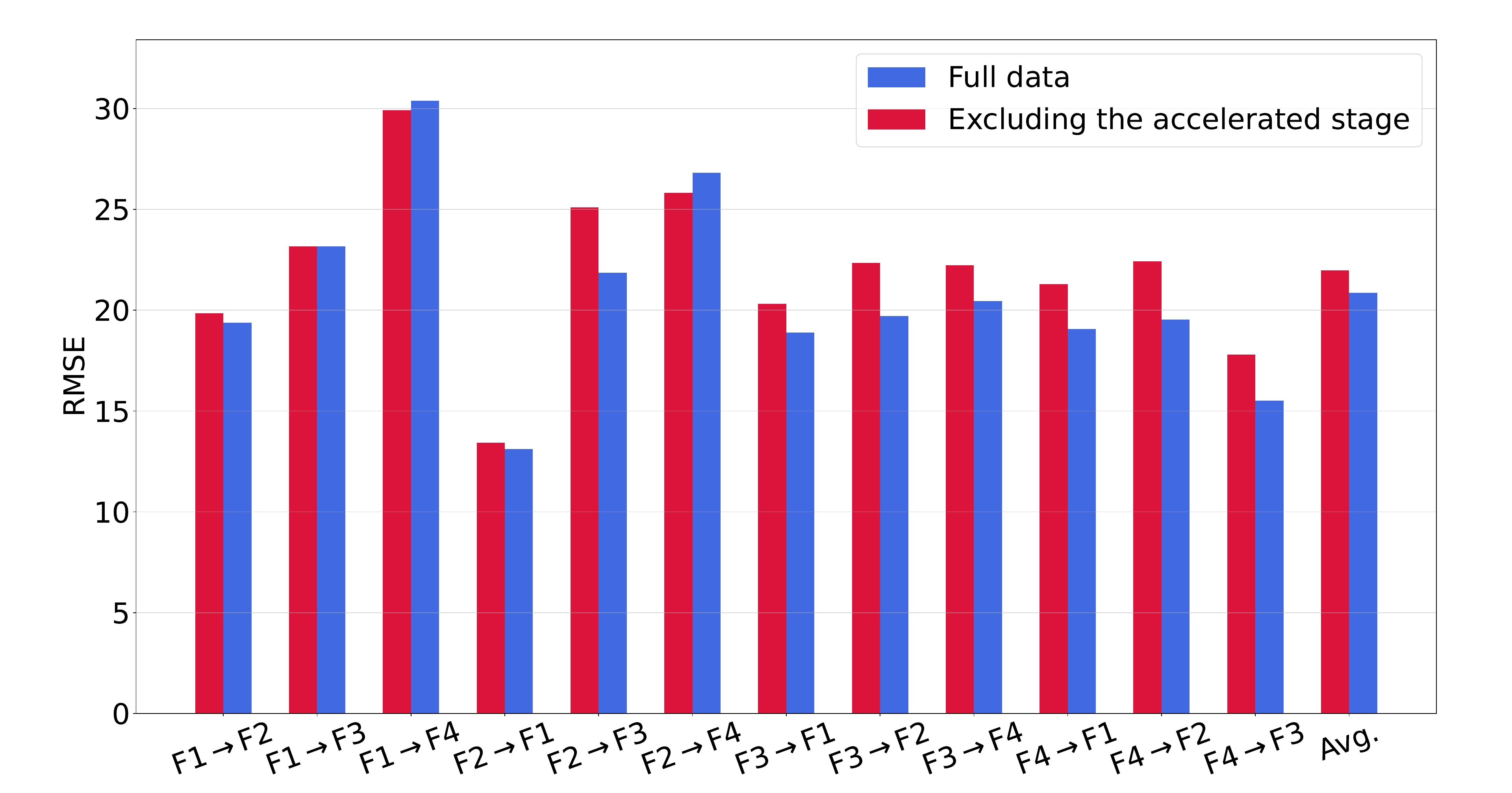}
    \includegraphics[width=0.45\textwidth]{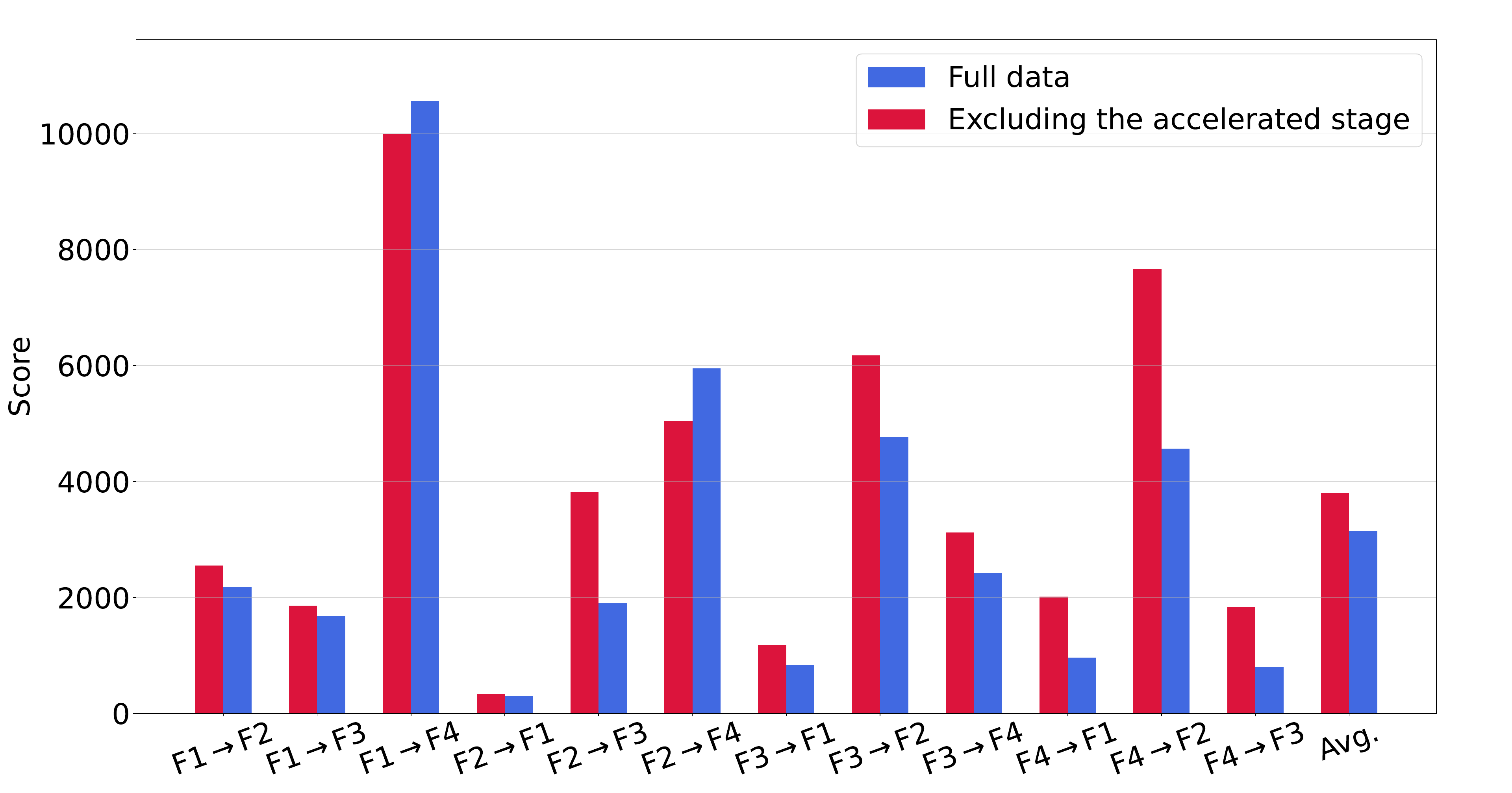}

    \caption{The insufficient degradation data analysis for excluding the accelerated degradation stage from the training dataset.}
    \label{fig:data_sensitivity}
\end{figure}

In practical real-world applications, acquiring data in the failure condition from target domain for training purposes poses challenges, which is prone to negative adaptation if match the whole source domain to the target domain. To examine the impact of insufficient degradation data for our proposed TACDA, we conducted an analysis. Specifically, we performed experiments by excluding data with accelerated degradation stage, from the dataset during training for domain adaptation. The results, depicted in Fig.~\ref{fig:data_sensitivity}, highlight the stability of the proposed method across various cross-domain scenarios. The rationale for this lies in the fact that degradation alignment exclusively aligns the distributions of source and target data in the shared degradation stage, thereby mitigating the influence stemming from inadequate degradation data within the target domain, such as poor alignment due to align data of accelerated stage to the sluggish stage.

\subsection{Analysis of Time Utilization and Resource Consumption}
We conducted additional experiments to analyze the temporal efficiency and resource consumption of our proposed method under four scenarios, and compared with the second-best method, i.e., DARE-DRAM. To ensure fair comparison, both TACDA and DARE-DRAM are trained for 180 epochs. As demonstrated in Table~\ref{tab:traing_time}, our method trains faster in two scenarios and is marginally slower in the other two. 
And TACDA requires a 10\% more GPU memory over DARE-DRAM. This increased memory usage is primarily due to the additional parameters and computations introduced by the decoder and discriminator, which allows the model to preserve target-specific information during adaption.
These results demonstrate that TACDA delivers superior performance without significantly higher computational costs, maintaining practical deployability while advancing state-of-the-art results.

\setcounter{table}{3}
\begin{table}[h]
\centering
\caption{Comparison of Time Utilization and Resource Consumption}
\label{tab:traing_time}
\begin{tabular}{l|l|c|c}
\toprule
\multicolumn{2}{c|}{Method} & TACDA & \makecell{DARE\\-GRAM} \\
\midrule
\multirow{4}{*}{\makecell{Time\\(second)}} & F1$\rightarrow$F2 & 979.4 & 784.8 \\
& F1$\rightarrow$F3 & 571.8 & 482.9 \\
& F2$\rightarrow$F3 & 575.0 & 1011.6 \\
& F2$\rightarrow$F4 & 1127.5 & 1141.2 \\
\midrule
\multicolumn{2}{c|}{GPU Memory (MiB)}  & 1792 & 1629 \\
\bottomrule
\end{tabular}
\end{table}
\section{Conclusion}
\label{conclusion}

In this paper, we introduced a novel approach called TACDA for the automatic discovery of domain-invariant features, while simultaneously preserving target-specific information and aligning the data within the same degradation stage in RUL prediction. The TACDA method was developed based on an adversarial auto-encoder domain adaptation framework, incorporating a soft-DTW loss and a data selection mechanism utilizing soft-DTW based K-means clustering and variance. To evaluate the effectiveness of the proposed TACDA method, extensive experiments were conducted, including a comprehensive comparison with various state-of-the-art domain adaptation techniques. The experimental results demonstrated that the TACDA method outperforms all other state-of-the-art methods significantly. Additionally, an ablation study was conducted to demonstrate the effectiveness of the key components in the domain adaptation process.

\bibliographystyle{IEEEtran}
\bibliography{references}
\end{document}